\documentclass[10pt,twocolumn,letterpaper]{article}

\usepackage[pagenumbers]{cvpr} 

\usepackage{microtype}

\renewcommand{\paragraph}[1]{\vspace{.5em}\noindent\textbf{#1.}}

\definecolor{cvprblue}{rgb}{0.21,0.49,0.74}
\usepackage[pagebackref,breaklinks,colorlinks,allcolors=cvprblue]{hyperref}
\usepackage{multirow}
\usepackage{booktabs}
\usepackage{makecell}
\usepackage[table]{xcolor}   
\usepackage{colortbl}        
\usepackage{cuted}      
\usepackage{caption}    
\usepackage[dvipsnames]{xcolor}

\newcolumntype{R}{>{\columncolor{red!5}\centering\arraybackslash}c}
\newcolumntype{B}{>{\columncolor{blue!5}\centering\arraybackslash}c}

\def\paperID{25198} 
\def\confName{CVPR}
\def\confYear{2026}

\title{Failure Modes for Deep Learning–Based Online Mapping: \\ How to Measure and Address Them}

\author{Michael Hubbertz\textsuperscript{1,2}\\
\and
Qi Han\textsuperscript{2}\\
\and
Tobias Meisen\textsuperscript{1}\\
\and
    \centerline{
        \textsuperscript{1}\textit{University of Wuppertal}, {\tt <lastname>@uni-wuppertal.de}
    } \\
    \centerline{
        \textsuperscript{2}\textit{Aptiv Services Deutschland GmbH}, {\tt <firstname>.<lastname>@aptiv.com}
    }
}

\begin{document}
\maketitle
\begin{abstract}

Deep learning-based online mapping has emerged as a cornerstone of autonomous driving, yet these models frequently fail to generalize beyond familiar environments. We propose a framework to identify and measure the underlying failure modes by disentangling two effects: Memorization of input features and overfitting to known map geometries. We propose measures based on evaluation subsets that control for geographical proximity and geometric similarity between training and validation scenes. We introduce Fréchet distance–based reconstruction statistics that capture per-element shape fidelity without threshold tuning, and define complementary failure-mode scores: a localization overfitting score quantifying the performance drop when geographic cues disappear, and a map geometry overfitting score measuring degradation as scenes become geometrically novel. Beyond models, we analyze dataset biases and contribute map geometry-aware diagnostics: A minimum-spanning-tree (MST) diversity measure for training sets and a symmetric coverage measure to quantify geometric similarity between splits. Leveraging these, we formulate an MST-based sparsification strategy that reduces redundancy and improves balancing and performance while shrinking training size. Experiments on nuScenes and Argoverse 2 across multiple state-of-the-art models yield more trustworthy assessment of generalization and show that map geometry-diverse and balanced training sets lead to improved performance. Our results motivate failure-mode-aware protocols and map geometry-centric dataset design for deployable online mapping. \href{https://github.com/mhubbertz/online_mapping_geometric_splits}{GitHub Page}

\end{abstract}    
\section{Introduction}
\label{sec:intro}

Maps are essential navigation tools, offering critical spatial information that helps drivers safely and efficiently navigate intricate road systems. Standard definition (SD) maps typically provide enough detail for human drivers to understand general road layouts. However, high-definition (HD) maps have become vital for autonomous vehicles, delivering detailed information including precise lane structures which complement onboard sensor perception of the dynamic environment and ensure safe navigation.

While traditional HD maps are constructed offline, autonomous driving systems require the ability to generate and update maps in real-time as the vehicle traverses new environments. This task, commonly referred to as \emph{online mapping}, leverages multimodal sensor inputs such as lidar, radar, and camera data. Contemporary research has shifted toward deep learning-based methods, as conventional rule-based approaches are unable to capture the complex spatial and semantic relationships necessary for accurate HD map generation.  

\begin{figure}[t] 
\includegraphics[width=0.45\textwidth]{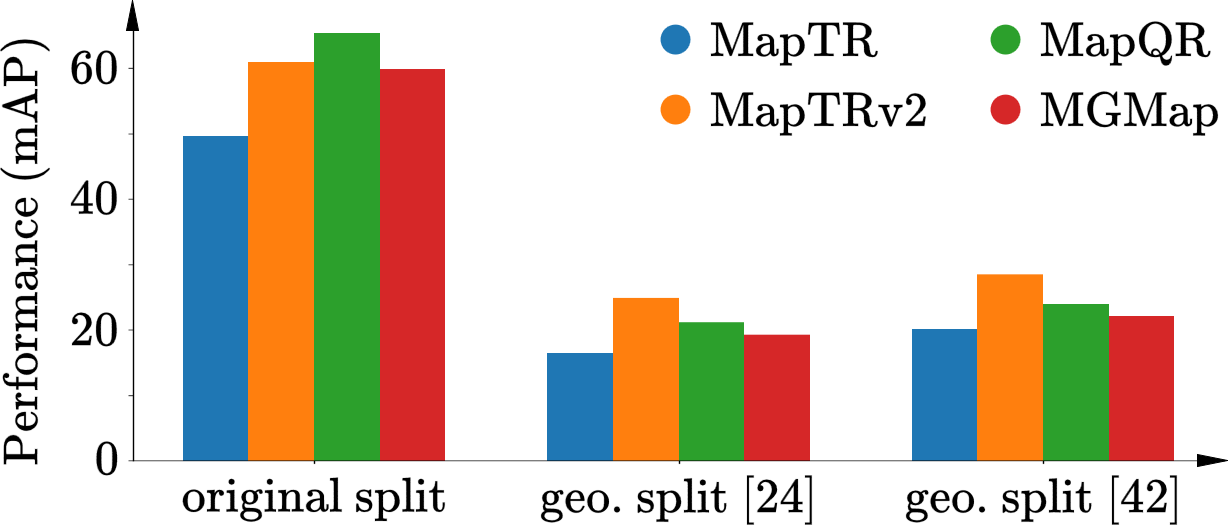}
\caption{Validation set performance of different state-of-the-art online mapping models on the nuScenes dataset \cite{caesar_nuscenes_2020} with original and two different geographically disjoint \cite{lilja_localization_2024, yuan_streammapnet_2024} splits.}
\label{fig:motivation}
\end{figure}


Despite remarkable progress, deep learning-based online mapping models still suffer from significant \emph{failure modes}. Lilja et al. have highlighted that models trained on widely used datasets such as nuScenes and Argoverse~2 exhibit strong geographic memorization effects rather than true generalization~\cite{lilja_localization_2024}. When evaluated on geographically overlapping splits, models show inflated performance, whereas performance drops drastically under geographically disjoint splits \cite{lilja_localization_2024, yuan_streammapnet_2024, qin_unifusion_2023, roddick_predicting_2020} (cf. \cref{fig:motivation}). Furthermore, dataset biases amplify the observed generalization issues through skewed training and evaluation, as the data may lack sufficient diversity in road structures and layouts.

In this work, we aim to examine these model and dataset related failure modes in detail. We propose new evaluation criteria to disentangle model memorization from generalization and provide dataset analysis and modification tools to quantify and minimize geometric biases.

Our key contributions can be summarized as follows:
\begin{itemize}
    \item We propose new evaluation measures based on discrete Fréchet distance that better capture geometric reconstruction quality compared to traditional average precision metrics based on Chamfer distance.
    \item We introduce a systematic framework to categorize and measure model failure modes in online mapping, disentangling localization overfitting from geometric overfitting.
    \item We conduct extensive experiments across multiple state-of-the-art online mapping models, examining their failure modes in more detail.
    \item We analyze biases in online mapping datasets, introducing new measures of geometric diversity and geometric similarity between splits and propose new splits with higher dissimilarity regarding map geometries.
    \item We propose a minimum spanning tree-based dataset sparsification strategy that reduces redundancy, enhances geometric balancing, and improves model generalization.
\end{itemize}

\section{Related Work}
\label{sec:related_work}

\subsection{Online Mapping Model Architectures}

HDMapNet \cite{li_hdmapnet_2022} pioneered deep learning-based online mapping as the first method to directly generate vectorized map elements from sensor data. Building on this, more recent approaches adopt Detection Transformer (DETR) inspired designs \cite{carion_end--end_2020}, where sensor features are projected into bird’s-eye view space and processed by transformer decoders with learnable object queries, followed by classification and regression heads to identify map categories and shapes. Key examples include MapTR \cite{liao_maptr_2023}, VectorMapNet \cite{liu_vectormapnet_2023}, and MapTRv2 \cite{liao_maptrv2_2023}.

Further enhancements have optimized this foundational architecture. Examples include incorporating mask guidance to improve map object predictions \cite{liu_mgmap_2024}, refining query designs within transformer decoders \cite{liu_leveraging_2024}, integrating prior and additional information \cite{xiong_neural_2023, li_bi-mapper_2023, li_enhancing_2024, jiang_p-mapnet_2024, wang_priormapnet_2024, song_memfusionmap_2024, sun_mind_2024}, and employing temporal fusion techniques to mitigate inconsistencies caused by occlusions and complex environments \cite{li_dtclmapper_2024, yuan_streammapnet_2024, chen_maptracker_2024, peng_prevpredmap_2024}.

Building on the generalization challenges outlined in \cref{sec:intro}, we employ several state-of-the-art online mapping architectures as experimental testbeds to systematically analyze failure modes. We deliberately include diverse architectures to reduce model-specific bias and to draw conclusions about localization and map geometry overfitting that are robust across different design choices.

\subsection{Overfitting Countermeasures and Metrics}

Deep learning-based online mapping models are particularly prone to overfitting. Lilja et al. demonstrated in their study that rather than learning generalizable structures, they tend to memorize location-specific patterns during training \cite{lilja_localization_2024}. This highlights a form of structural memorization that differs from conventional overfitting. Consequently, standard countermeasures such as early stopping, weight decay, and dropout \cite{dietterich_overfitting_1995, hawkins_problem_2004, srivastava_dropout_2014} are insufficient, motivating the need for more domain-specific strategies.

To measure how much a deep learning model overfits, Aburass and Rumman \cite{aburass_quantifying_2024} introduced the \emph{Overfitting Index}, combining accuracy and loss to track generalization across training epochs. Margin-based distributions can also predict the generalization gap \cite{huang_practical_2025, jiang_predicting_2019, jiang_neurips_2020}, while reviews by Valle-Pérez and Louis \cite{valle-perez_generalization_2020} and critiques by Gastpar et al. \cite{gastpar_fantastic_2023} draw attention to the theoretical limits of generalization bounds.

In this work, we aim to investigate and quantify overfitting specifically for deep learning-based online mapping models. We aim to provide deeper insights into the core problems that lead to performance decrease for geographically disjoint splits, quantifying intuitive expectations and proposing improved performance scores and additional evaluation measures. We also propose countermeasures through pruning samples with redundant map geometries in training data, leading to improved balance during training.

\subsection{Online Mapping Datasets and Biases} \label{sec:related_work_datasets}

Recent advances in online mapping for automated driving have predominantly relied on supervised learning, requiring datasets that pair sensor inputs capturing the vehicle’s environment with corresponding HD ground-truth maps. Publicly available datasets that satisfy this requirement are Argoverse 1 and 2 \cite{chang_argoverse_2019, wilson_argoverse_2023}, nuScenes \cite{caesar_nuscenes_2020}, and Waymo \cite{sun_scalability_2020}, with nuScenes and Argoverse~2 emerging as predominantly used datasets in the domain of online mapping \cite{li_hdmapnet_2022, liao_maptr_2023, liu_vectormapnet_2023, liao_maptrv2_2023, liu_leveraging_2024, xiong_neural_2023, li_bi-mapper_2023, li_enhancing_2024, jiang_p-mapnet_2024, wang_priormapnet_2024, song_memfusionmap_2024, sun_mind_2024, li_dtclmapper_2024, yuan_streammapnet_2024, chen_maptracker_2024, peng_prevpredmap_2024}. Both datasets provide predefined splits that partition the data into training and validation sets. Thereby, inherent biases in these datasets and splits can skew training and evaluation, leading to inflated benchmark performance while hindering real-world generalization.

For example, Koe et al. and Marathe et al. showed that autonomous-driving datasets are biased toward clear-weather conditions, leading to poor robustness under adverse weather \cite{marathe_rain_2023, kou_generalizable_2025}. Similar issues have been explored in object detection, including studies on class imbalance \cite{katare_bias_2022, peri_towards_2022} and fairness in pedestrian detection  \cite{fernandez_llorca_attribute_2024, li_bias_2025, rajan_distribution-aware_2024}.

Research on dataset biases in online mapping has focused on geographic bias for dataset splits \cite{roddick_predicting_2020, qin_unifusion_2023, lilja_localization_2024, yuan_streammapnet_2024}. Geographic bias occurs when training and evaluation data come from overlapping regions, allowing models to rely on location-specific cues rather than learn generalizable representations, thereby inflating benchmark performance.

The original nuScenes and Argoverse~2 splits were originally designed for object detection as well as motion forecasting tasks and follow a temporal division, resulting in partially large geographical overlap \cite{lilja_localization_2024}. For the original nuScenes and Argoverse~2 splits, approximately 80 \% and 45 \% of validation samples, respectively, are within five meters of a sample used during training. To mitigate such a geographic bias, several geographical splits that minimize overlap for nuScenes \cite{roddick_predicting_2020, qin_unifusion_2023, lilja_localization_2024, yuan_streammapnet_2024} and Argoverse~2 \cite{lilja_localization_2024, yuan_streammapnet_2024} have been proposed recently.

Building on these observations, we revisit dataset bias for online mapping from a broader perspective. Beyond the well-studied issue of geographical sampling bias, we identify geometric similarity between train and validation scenes as a critical yet underexplored factor. We further quantify geometric diversity and similarity across existing splits and show their correlation with model performance, motivating the design of map geometry-aware splits for more reliable evaluation of generalization.
\section{Failure Modes for Online Mapping Models}

We hypothesize that performance drops between geographically overlapping and disjoint dataset splits reveal two recurring issues: Reliance on memorized input features (\emph{localization overfitting}) and overfitting to known map geometries (\emph{geometric overfitting}). To study these effects systematically, we first derive evaluation sets from the dataset splits that enable inference of models’ tendencies toward distinct modes of overfitting. Second, we introduce improved performance measures. Based on both, we introduce measures that systematically capture and quantify both model-based failure modes. In the following, the training and validation sets are denoted as $T$ and $V$, respectively, and the set of map object classes is defined as $C_{\text{map}}$.

\subsection{Evaluation Set Derivation} \label{eval_set_derivation}

As a first step, we introduce two criteria to further specify each validation sample. The first criterion is the geographical distance $d$ from each validation sample to its closest training sample, capturing potential leakage due to spatial overlap and concrete feature memorization. 

\begin{equation}
    d(v) := \min_{t \in T} \mathrm{dist}(v, t)
\end{equation}

where $\mathrm{dist}(v, t)$ denotes the geographical Euclidean distance (or a different suitable distance measure) between the validation sample $v \in V$ and the training sample $t \in T$.

However, two scenes may be geographically distant yet geometrically identical, or adjacent yet structurally novel. Consequently, we also assess how closely the map geometries align across different scenes. Therefore, we introduce the geometric similarity $s$ between the ground truth map of each validation sample and its most geometrically similar training sample.

\begin{equation}
    s(v) := \min_{t \in T} \mathrm{sim}(v, t)
\end{equation}

where $\mathrm{sim}(v, t)$ denotes the geometric similarity cost (low = similar, high = dissimilar) between the validation sample $v \in V$ and the training sample $t \in T$. 

For computing the geometry similarity cost $\mathrm{sim}(v,t)$, each class $c_i \in C_{\text{map}}$ is handled separately. We compare the two samples’ map elements $\mathcal{M}_v$ and $\mathcal{M}_t$ by forming the assignment-cost matrix $A_{v,t} \in \mathbb{R}^{|\mathcal{M}_v| \times |\mathcal{M}_t|}$. The cost between two elements (each a polygon or polyline with a variable number of points) is the minimum discrete Fréchet distance \cite{eiter_computing_1994} over all point orderings: For polylines, original and reversed; for polygons, all cyclic permutations in both orientations (original and reversed). A mathematical definition and details on computing the discrete Fréchet distance are provided in the supplementary material. Element assignments are obtained via minimum-cost bipartite matching on $A_{v,t}$. Let $a_{\text{matched}}$ be the sum of costs for matched pairs across classes, $n_{\text{matched}}$ their count, and $n_{\text{unmatched}}$ the number of unmatched elements, each incurring a fixed penalty $\delta$. The geometry similarity cost is then defined as 

\begin{equation}
\mathrm{sim}(v,t) := \frac{a_{\text{matched}} + n_{\text{unmatched}} \cdot \delta}{n_{\text{matched}} + n_{\text{unmatched}}}.
\end{equation}

Normalizing by $n_{\text{matched}} + n_{\text{unmatched}}$ prevents samples with few elements from being overemphasized.

We provide an instructive visualization of $\mathrm{sim}(v,t)$ for an exemplary sample pair in the supplementary material. To check whether the inductive biases of current DETR-based mapping decoders correlates more with rotation- and translation-invariant topological features rather than geometrical features, we also provide an ablation study with a topological similarity measure in the supplementary material.

\begin{figure*}[t] 
\centering
\includegraphics[width=0.9\textwidth]{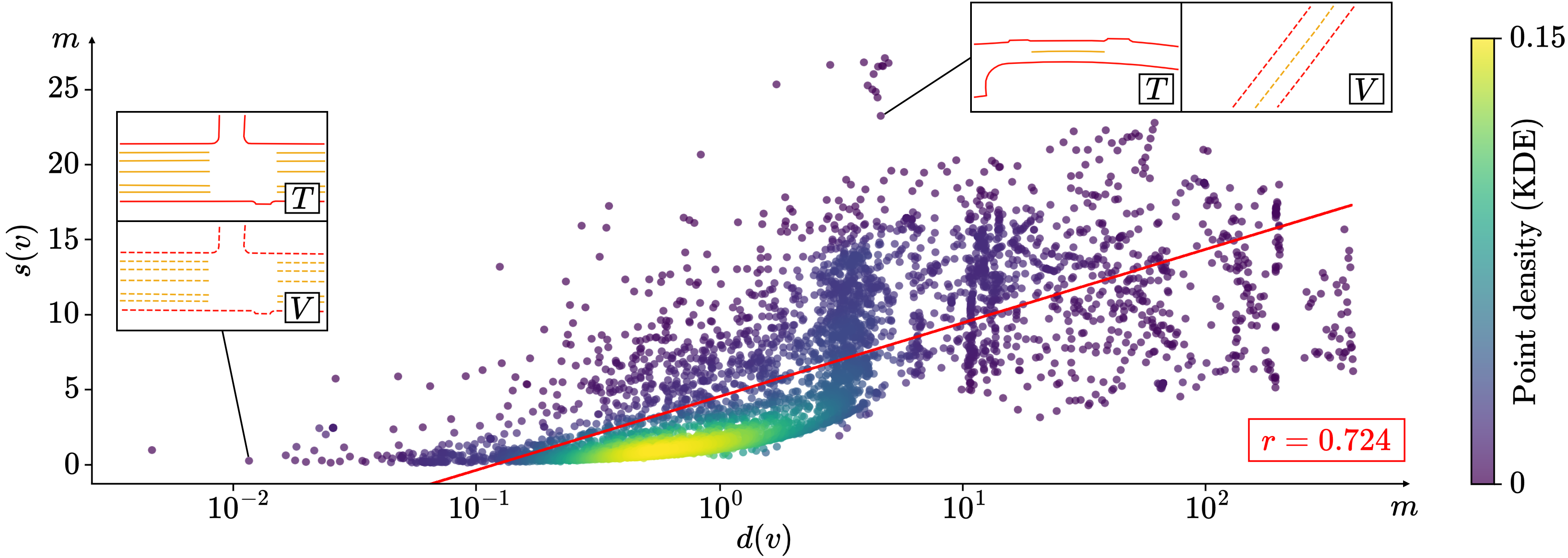}
\caption{Correlation between $d(v)$ and $s(v)$ for the nuScenes original split (Pearson correlation coefficient $r = 0.724$). Three representative pairs of $s(v)$ are presented, showing each validation sample alongside its closest geometric match from the training set.}
\label{fig:d_s_correlation}
\end{figure*}

As we intend to derive disentangled failure mode measures based on $s(v)$ and $d(v)$ later on, it is important to examine how these two quantities correlate. As seen in \cref{fig:d_s_correlation}, both measures are strongly positively correlated on the nuScenes original split: Validation samples in close proximity to the training set tend to have more similar map geometries (as expected), whereas more distant samples generally exhibit more dissimilar map geometries.

Next, we can separate the validation split $V$ into sample sets that provide information about the model failure modes. First, we divide $V$ into two subsets $V_\text{close},\:V_\text{far} \subset V$ based on $d(v)$ with the distance threshold $T_\text{dist}$:

\begin{align} 
    V_{\text{close}} &:= \{ v \in V \mid d(v) \leq T_{\text{dist}} \}, \\
    V_{\text{far}} &:= \{ v \in V \mid d(v) > T_{\text{dist}} \}.
\end{align}

The separation indicates whether the deep learning model has already been trained with a sample in close vicinity to the validation sample or not and would be able to utilize knowledge about memorized location-specific features. We therefore expect the performance of validation samples in $V_{\text{close}}$ to be superior compared to $V_{\text{far}}$.

However, to draw valid conclusions about the performance drop between $V_\text{close}$ and $V_\text{far}$, one must also account for $s(v)$, given its strong correlation with $d(v)$ and to avoid conflating localization overfitting with geometric overfitting. 

To quantify localization overfitting, we refine the $V_\text{close}$ and $V_\text{far}$ sets by alignment of the distributions of $s(v)$ across both sets. The empirical cumulative distribution $F_{s,V_\text{sub}}$ of $s(v_\text{sub})$ for a subset $V_\text{sub} \subset V$ is described by

\begin{equation}
    F_{s,V_{\text{sub}}}(t)\;:=\;\frac{1}{|V_\text{sub}|}\,\bigl|\{\,v_\text{sub}\in V_\text{sub} \mid s(v_\text{sub})\le t\}\bigr| .  
\end{equation}

To match the distributions w.r.t. $s(v)$ between $V_\text{close}$ and $V_\text{far}$, we sample the subsets $V_\text{close*} \subset V_\text{close}$ and $V_\text{far*} \subset V_\text{far}$ by approximating 

\begin{equation}
    F_{s,V_\text{close*}}(t) \approx F_{s,V_\text{far*}}(t)\;\forall\,t.
\end{equation}

To obtain $V_\text{close*}$ and $V_\text{far*}$, we perform bipartite matching of $s(v)$ across both sets, followed by filtering based on a predefined threshold for the absolute value of the differences between $s(v)$ for the matched pairs. To ensure that the geometric similarity distributions of $s(v)$ for $V_\text{close}$ and $V_\text{far*}$ are closely aligned, we choose the threshold such that their Kullback–Leibler divergence remains below 0.01.

We further examine geometric overfitting by stratifying the geographically distant subset $V_{\text{far}}$. 
Let $\tau_0 \le \tau_1 \le \ldots \le \tau_b$ partition $[\min_{v\in V_{\text{far}}} s(v),\, \max_{v\in V_{\text{far}}} s(v)]$ into $b$ equal-width intervals. For $i=1,\dots,b$, we define

\begin{equation}
B_i := \{\, v \in V_{\text{far}} \mid \tau_{i-1} < s(v) \le \tau_i \,\}.
\end{equation}

Because $V_{\text{far}}$ enforces no geographic overlap with training data, variation across $\{B_i\}_{i=1}^b$ isolates the effect of map geometry alone: Low-index bins (low $s(v)$) group geometrically similar scenes, whereas high-index bins (high $s(v)$) group geometrically dissimilar scenes. Evaluating the performance per bin yields a geometry–performance curve that quantifies sensitivity to geometric novelty.

\subsection{Performance Scores}\label{sec:performance_scores}

To introduce measurements for the previously defined failure modes, we first aim to quantify the performance of the examined online mapping model on the evaluation sets.

To ensure comparability with recent online mapping works \cite{li_hdmapnet_2022, liao_maptr_2023, liu_vectormapnet_2023, liao_maptrv2_2023, liu_leveraging_2024, xiong_neural_2023, li_enhancing_2024, jiang_p-mapnet_2024, wang_priormapnet_2024, song_memfusionmap_2024, sun_mind_2024, li_dtclmapper_2024, yuan_streammapnet_2024, chen_maptracker_2024, peng_prevpredmap_2024} we adopt the standard average precision (AP) metric with the Chamfer distance as the matching criterion, details and definitions are provided in the supplementary material. For each class $c_i \in C_{\text{map}}$, we compute the AP across all samples in the set by declaring a prediction a true positive if its Chamfer distance $D_{\text{Chamfer}}$ to the ground truth is below a threshold $\tau$. We evaluate the AP separately for each threshold $\tau \in T$, where $T = \{0.5, 1.0, 1.5\}$, and then take the average across all thresholds $T$ and all classes $C_{\text{map}}$ to obtain the final mean average precision (mAP) score used for comparison:

\begin{equation}
    \text{mAP} = \frac{1}{|C_{\text{map}}||T|} \sum_{c_i \in C_{\text{map}}} \sum_{\tau \in T} \text{AP}_{c_i,\tau}.
\end{equation}

While the Chamfer distance-based AP suffices for performance estimates for large sets of samples, it struggles to capture the performance for sets with low amounts of samples due to its sensitivity to discrete outcomes and lack of granularity in small sample scenarios. Because matches are discretized into hits or misses, the metric can be sensitive to individual outcomes and lacks granularity regarding \emph{how close} a near-miss is, failing to capture the exact degree of similarity between prediction and ground truth. Moreover, the Chamfer distance is permutation-invariant and only enforces proximity of points, not their global arrangement, which limits its ability to assess whether the reconstructed map element preserves the correct shape. 


\begin{figure}[t] 
\includegraphics[width=0.48\textwidth]{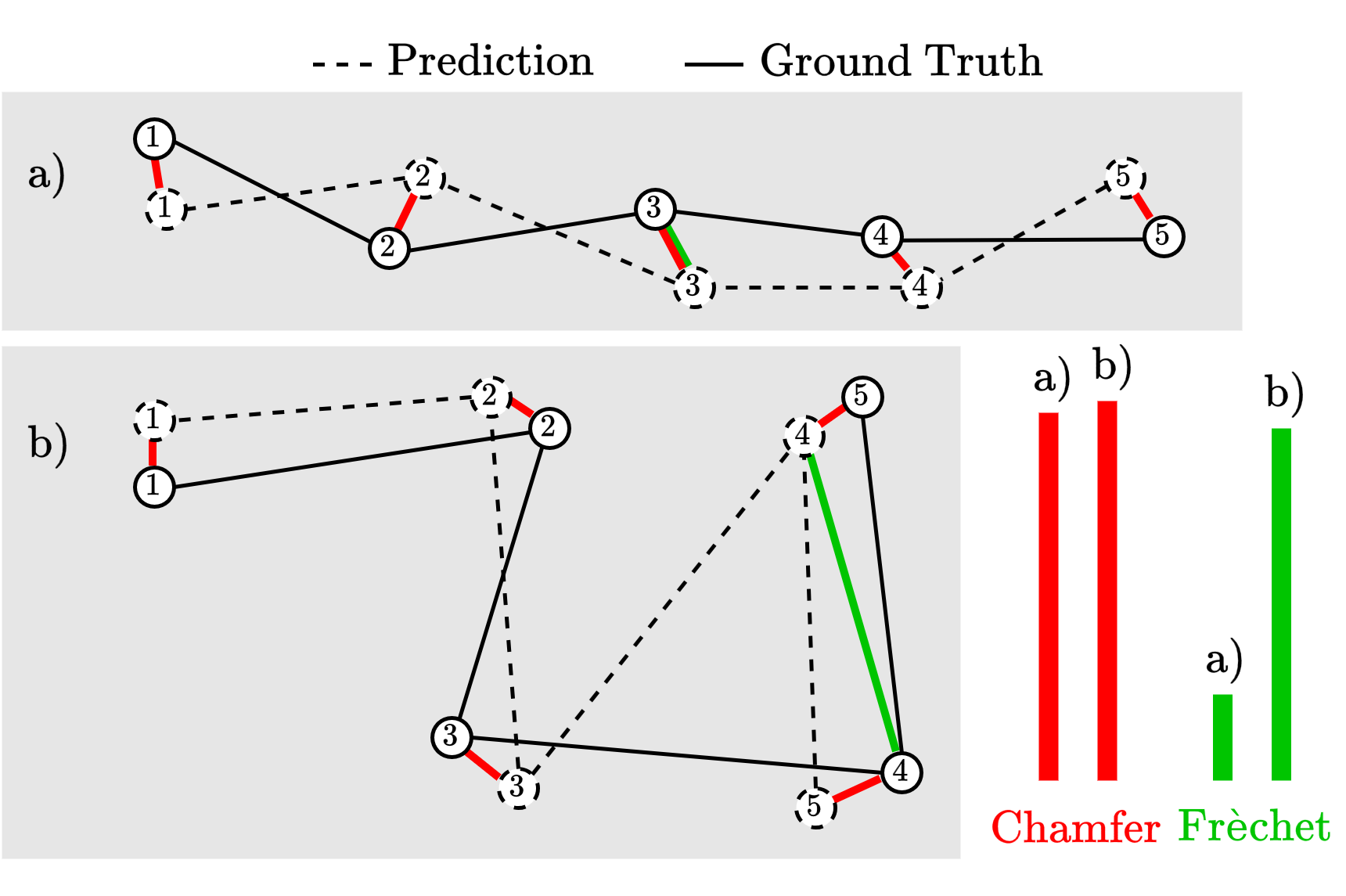}
\caption{Visualization of two exemplary prediction and ground truth map element pairs. The \textcolor{red}{Chamfer distance (red)} and \textcolor{ForestGreen}{Fréchet distance (green)} are shown as performance metrics for both cases. In example (a), both metrics produce meaningful results. In example (b), the Chamfer distance remains nearly unchanged compared to (a) because it ignores point ordering, whereas the Fréchet distance yields a much higher value, capturing the larger geometric deviation.}
\label{fig:chamfer_frechet_comparison}
\end{figure}

To address these limitations, we introduce an intuitive measure of the quality of ground-truth map geometry reconstruction. This new performance measure is based on the discrete Fréchet distance. It takes into account the order of points within each map element, while being independent of their individual lengths. In contrast to the Chamfer distance, this yields a more accurate assessment of per-element similarity (cf. \cref{fig:chamfer_frechet_comparison}). Analogous to $\mathrm{sim}(\cdot,\cdot)$, we compute class-wise bipartite matching between the prediction $P$ and the ground truth $G$ and collect the matched costs into a distribution $D$. To assess how well the prediction reconstructs the ground-truth map, we use the median $M$ and the interquartile range $IQR$:

\begin{equation}
M = \operatorname{median}(D) \quad \text{and} \quad
IQR = Q_3(D) - Q_1(D)
\end{equation}

where $Q_1(D)$ and $Q_3(D)$ denote the first and third quartiles of $D$. Unlike mAP, they avoid threshold calibration and remain informative for sets of varying sizes, being less sensitive to discrete outcomes and enabling comparisons at the single-sample level. Additional details for $M$ can be found in the supplementary material.

\subsection{Measures for Failure Modes} \label{model_failure_mode_metrics}

As stated in \cref{eval_set_derivation}, the performance drop between $V_\text{close*}$ and $V_\text{far*}$ describes the model's extent of localization overfitting. We utilize the proposed Fréchet distance-based measure $M$ across all classes for quantifying the performance drop, to improve the significance of the results for sets with low amounts of samples compared to mAP. We propose the following localization overfitting score $\mathcal{O}_\text{loc}$ for measuring this property of a model:

\begin{equation}
    \mathcal{O}_\text{loc} := \frac{M_\text{far*} - M_\text{close*}}{M_\text{close*}}
\end{equation}

This score thus measures the relative performance drop. A value near zero indicates that the model generalizes well to novel feature combinations ($M_\text{far*} \approx M_\text{close*}$), whereas a large value reveals a strong reliance on memorizing concrete location-specific features ($M_\text{far*} \gg M_\text{close*}$).

To quantify geometric overfitting, we measure the performance decay of $M_{\text{far},i}$ across bins $B_i$. This is estimated via linear regression, where the slope of the fitted line represents the rate of performance drop. To enable an arbitrary number of bins $b$, we use the mean value of $s(v)$ within each interval $[\tau_{i-1}, \tau_i]$, denoted as $\mu_{\text{s,far,i}}$, together with the corresponding $M_{\text{far},i}$ value for the linear regression. To account for differing bin sizes, we weight the regression data points by their sample counts $w_i$. Formally, we propose the following geometry overfitting score $\mathcal{O}_\text{geom}$:

\begin{equation}
    \begin{split}
        &\mathcal{O}_\text{geom}
        =
        \frac{\sum_{i=1}^{b} p_i\bigl(\mu_{\text{s,far},i}-\mu_x\bigr)\bigl(M_{\text{far},i}-\mu_y\bigr)}
        {\sum_{i=1}^{b} p_i\bigl(\mu_{\text{s,far},i}-\mu_x\bigr)^2} \quad \text{where} \\[4pt]
        &p_i = \frac{w_i}{\sum_{j=1}^b w_j}, \ \:
        \mu_x = \sum_{i=1}^b p_i\,\mu_{\text{s,far},i}, \ \:
        \mu_y = \sum_{i=1}^b p_i\,M_{\text{far},i}.
    \end{split}
\end{equation}

A value close to zero for $\mathcal{O}_\text{geom}$ indicates great generalization properties regarding map geometry ($M_{\text{far},1}~\approx~M_{\text{far},b}$). The higher the value of $\mathcal{O}_\text{geom}$, the more the performance decreases across bins and the model tends to overfit on the learned training map geometries.

\section{Dataset Biases in Online Mapping} \label{sec:dataset_biases}

Beyond model-specific failure modes, dataset bias can distort training and evaluation in online mapping. Building on \cref{model_failure_mode_metrics}, we consider two relevant biases: \emph{geographical bias} (spurious performance improvements from geographical overlap and resulting feature memorization from $T$ to $V$) and \emph{geometric bias} (performance gains from the similarity of map geometries across $T$ and $V$).

Geographical bias in widely used online mapping datasets such as nuScenes \cite{caesar_nuscenes_2020} and Argoverse 2 \cite{wilson_argoverse_2023} has been extensively studied, as discussed, and geographically disjoint splits are available to mitigate this effect \cite{lilja_localization_2024, qin_unifusion_2023, roddick_predicting_2020, yuan_streammapnet_2024}. Building on these efforts, we turn our attention to geometric biases in these datasets, which remain largely unexplored.

\subsection{Geometric Diversity of Sets}

We hypothesize that online mapping models achieve better performance when the training data exhibits high geometric diversity and a well-balanced representation across different map geometries, as opposed to datasets dominated by a few or highly similar geometries. 

To confirm this, we introduce a dataset-level measure of geometric diversity. For a given sample set $D = \{s_1,...,s_M\}$, we construct a fully connected weighted graph $\mathcal{G}_\text{sim}$ where each node $n_i \in \mathcal{N}$ corresponds to a sample $s_i$ and each edge $e_{ij} \in \mathcal{E}$ is assigned the corresponding similarity cost $\mathrm{sim}(s_i,s_j)$. From $\mathcal{G}_\text{sim}$, we extract the minimum spanning tree (MST) $\mathcal{T}_\text{sim}$. 


We take the sum of all edge weights $e_{ij} = \mathrm{sim}(s_i,s_j)$ in $\mathcal{T}_\text{sim}$ as a measure for the geometric diversity of the sample set $D$ and define it as $\mathrm{geomdiv}(D)$:

\begin{equation}
    \mathrm{geomdiv}(D) := \sum_{(i,j) \in \mathcal{E}(\mathcal{T}_\text{sim})} \mathrm{sim}(s_i,s_j).
\end{equation}

High values of $\mathrm{geomdiv}(D)$ indicate a set with diverse and varied map geometries, while low values suggest redundancy and limited geometric variety. This approach offers an interpretable method for comparing splits, monitoring geometric coverage as data volume increases, and selecting subsets that maintain structural diversity.

\subsection{Geometric Similarity between Sets}

Besides geometric diversity, geometric similarity between training and validation splits could also influence evaluation performance. We expect prediction performance to improve when the map geometry distributions of the training and evaluation sets are similar, compared to cases where these distributions differ significantly. 

To capture geometric similarity between two sets $D_1$ and $D_2$, we define it as their symmetric coverage-based similarity. The directed cover $\mathrm{cov(D_1 \to D_2)}$ measures how well $D_2$ covers the map geometries in $D_1$ by averaging the nearest-neighbor map geometry similarity cost from each sample $s_{1,i}\in D_1$ into $D_2$:

\begin{equation}
    \mathrm{cov}(D_1 \!\to\! D_2)
      :=
      \frac{1}{|D_1|}
      \sum_{s_{1,i} \in D_1}
      \min_{s_{2,j} \in D_2} \,\mathrm{sim}(s_{1,i},s_{2,j}).
    \label{eq:cover}
\end{equation}

Because coverage is asymmetric, we define the geometric similarity between both sets $\mathrm{geomsim(D_1, D_2)}$ as the mean of both directions, yielding an order-agnostic score that attains low values only when the sets mutually cover each other’s map geometries:

\begin{equation}
    \mathrm{geomsim}(D_1,D_2)
      :=
      \tfrac{1}{2}\bigl(
        \mathrm{cov}(D_1 \!\to\! D_2)
        +
        \mathrm{cov}(D_2 \!\to\! D_1)
      \bigr).
    \label{eq:geomsim}
\end{equation}
\section{Experiments}
\label{sec:experiments}


\subsection{Experimental Setup}

\paragraph{Models}
Our primary analyses use MapTRv2~\cite{liao_maptrv2_2023} as a representative state-of-the-art online mapping architecture to validate the proposed evaluation protocol and failure-mode measures. To assess whether the observations generalize across design choices, we additionally examine a basic model (MapTR~\cite{liao_maptr_2023}) and more advanced variants with different transformer decoder query architecture (MapQR~\cite{liu_leveraging_2024}) and auxiliary masking task (MGMap~\cite{liu_mgmap_2024}). All models were trained by us based on the publicly available codebases and configurations.

\paragraph{Datasets and Splits}
We use the nuScenes and Argoverse~2 datasets. Following prior work, we compare the original dataset splits with geographically disjoint splits (near extrapolation split by \cite{lilja_localization_2024} and geographical split by \cite{ yuan_streammapnet_2024}). Furthermore, we derive geometric dataset splits for the whole dataset that maximize dissimilarity between training and evaluation splits. All splits partition the dataset into 70/15/15\% for the training, validation, and test sets, respectively. More details about the derivation of geometric dataset splits are provided in the supplementary material.

\subsection{Measure Verification and Results}

\begin{figure*}[ht] 
\centering
\includegraphics[width=\textwidth]{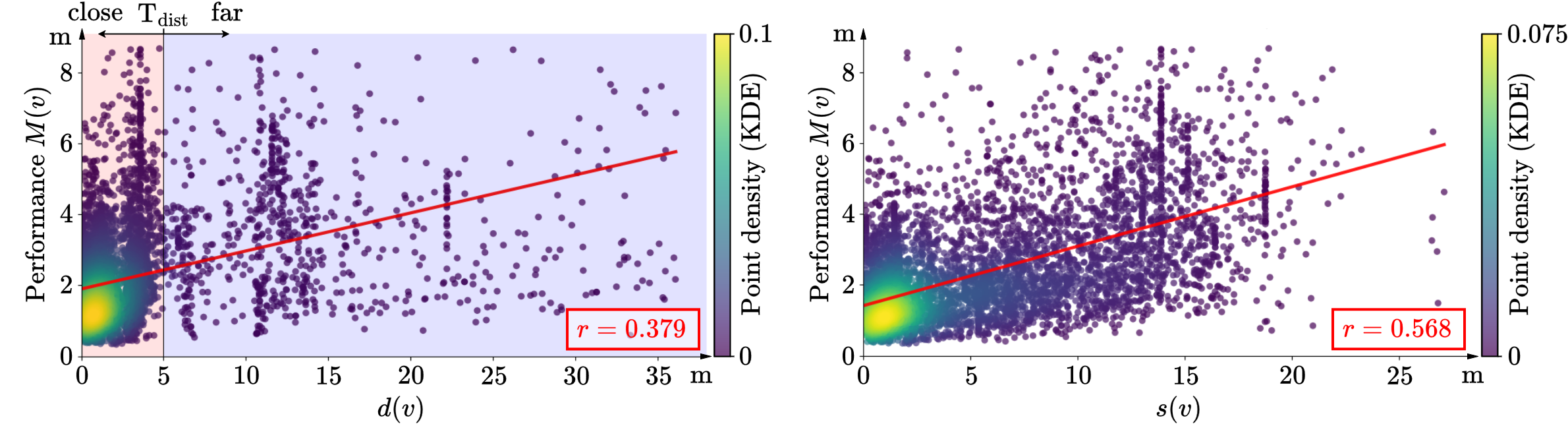}
\caption{Sample-wise performance of MapTRv2 measured by $M$ for the nuScenes original split, plotted over $d(v)$ (left) and $s(v)$ (right). In the left plot, an exemplary $T_\text{dist} = 5\,m$ is displayed, separating the validation set $V$ into $V_\text{close}$ and $V_\text{far}$. Performance is positively correlated with both $d(v)$ and $s(v)$, with the correlation being stronger for $s(v)$ (Pearson correlation coefficient $r = 0.568 > 0.379$).}
\label{fig:d_s_vs_M}
\end{figure*}

\begin{table*}[ht]
    \centering
    \begin{tabular}
        {%
            ll   
            cccccccc    
        }
        \multicolumn{2}{l}{Dataset and split} & $\mathrm{geomdiv}(T)$ & $\mathrm{geomdiv}(V)$ & $\mathrm{geomsim}(T, V)$ & $d(v) < 5\,m$ & $\text{mAP} \uparrow$ & $M^{\pm IQR} \downarrow$ & $\mathcal{O}_\text{loc} \downarrow$ & $\mathcal{O}_\text{geom} \downarrow$\\[0.3em]
        \toprule
        \multirow{4}{*}{\rotatebox{90}{nuScenes}} & original & 96.8 km & 30.1 km & 8.32 m & 79.47 \% & 60.95 & $1.94^{\pm 3.05}$ & 24.73 & 21.22 \\
        & geo. \cite{lilja_localization_2024} & 80.6 km & 22.5 km & 14.66 m & 0.95 \% & 24.96 & $4.07^{\pm 6.14}$ & n.a. & 9.75 \\
        & geo. \cite{yuan_streammapnet_2024} & 90.2 km & 22.4 km & 13.85 m & 0 \% & 28.53 & $3.24^{\pm 5.50}$ & n.a. & 13.84 \\
        & geometric & 91.3 km & 10.7 km & 21.08 m & 8.53 \% & 28.37 & $4.17^{\pm 6.08}$ & 4.40 & 10.49 \\
        \midrule
        \multirow{4}{*}{\rotatebox{90}{Argoverse 2}} & original* & 91.0 km & 23.7 km & 8.98 m & 44.89 \% & 63.97 & $1.77^{\pm 2.99}$ & 7.29 & 11.17 \\
        & geo. \cite{lilja_localization_2024}* & 87.3 km & 19.8 km & 11.37 m & 0 \% & 49.53 & $2.40^{\pm 4.18}$ & n.a. & 11.97 \\
        & geo. \cite{yuan_streammapnet_2024}* & 97.2 km & 18.6 km & 10.47 m & 0 \% & 57.61 & $2.13^{\pm 3.69}$ & n.a. & 11.15 \\
        & geometric* & 90.7 km & 18.6 km & 26.51 m & 4.73 \% & 34.13 & $2.59^{\pm 4.30}$ & 9.37 & 7.98 \\
        \bottomrule
    \end{tabular}
    \caption{Comparison of original, geographically disjoint \cite{lilja_localization_2024, yuan_streammapnet_2024} and geometric splits for nuScenes and Argoverse 2. Dataset split geometric diversities $(\mathrm{geomdiv}$) and similarities ($\mathrm{geomsim}$) are listed for training set $T$ and validation set $V$ as well as their geographical overlap ($d(v) < 5 \,m$). For the dataset properties, the Argoverse 2 splits (*) were subsampled from 10 Hz to 2 Hz to enable comparison with nuScenes. Performance and overfitting measures are provided for MapTRv2 \cite{liao_maptrv2_2023}.}
    \label{tab:split_comparison}
\end{table*}

\begin{table*}[ht]
    \centering
    \begin{tabular}
        {%
           l   
          RR  
          BBBBBB 
        }
         &
        \multicolumn{2}{>{ \columncolor{red!5} }c}{\parbox[b][1.0cm]{3cm}{\shortstack[b]{\textbf{Localization Overfitting} \\ \textbf{Score $\mathcal{O}_\text{loc}$ (\%) $\downarrow$}}}} & 
        \multicolumn{6}{>{ \columncolor{blue!5} }c}{\textbf{Geometric Overfitting Score $\mathcal{O}_\text{geom}$ (\%) $\downarrow$}} \\[0.65em]
         & 
        \parbox{1.8cm}{\centering \makecell{nuScenes}} & 
        \parbox{1.8cm}{\centering \makecell{Argoverse 2}} & 
        \multicolumn{3}{>{ \columncolor{blue!5} }c}{\text{nuScenes splits}} & 
        \multicolumn{3}{>{ \columncolor{blue!5} }c}{\text{Argoverse 2 splits}} \\[0.3em]
        \parbox[t][0.45cm]{1.2cm}{Model} & 
        original split & 
        original split & 
        {\parbox{1.2cm}{\centering \text{original}}} & 
        {\parbox{1.2cm}{\centering          \text{geo. \cite{lilja_localization_2024}}}} &
        {\parbox{1.2cm}{\centering          \text{geo. \cite{yuan_streammapnet_2024}}}} &
        {\parbox{1.2cm}{\centering \text{original}}} & 
        {\parbox{1.2cm}{\centering          \text{geo. \cite{lilja_localization_2024}}}} &
        {\parbox{1.2cm}{\centering          \text{geo. \cite{yuan_streammapnet_2024}}}}
        \\
        \specialrule{\lightrulewidth}{0pt}{0pt}
        
        \parbox[b][0.45cm]{2.2cm}{MapTR \cite{liao_maptr_2023}} & \textbf{24.42} & 22.06 & \textbf{18.66} & 13.55 & 18.66 & \textbf{10.69} & 22.43 & 13.86 \\
        MapTRv2 \cite{liao_maptrv2_2023} & 24.73 & \textbf{7.29} & 21.22 & \textbf{9.75} & \textbf{13.84} & 11.17 & 11.97 & 11.15 \\
        MapQR \cite{liu_leveraging_2024}& 57.07 & 13.91 & 21.03 & 10.87 & 19.48 & 10.73 & \textbf{11.37} & \textbf{9.08} \\
        MGMap \cite{liu_mgmap_2024} & 33.19 & 22.06 & 24.12 & 14.06 & 17.42 & \textbf{10.69} & 23.50 & 24.47 \\
    \end{tabular}
    \caption{$\mathcal{O}_\text{loc}$ and $\mathcal{O}_\text{geom}$ for different state-of-the-art online mapping models on nuScenes and Argoverse 2 datasets with original and geographically disjoint splits \cite{lilja_localization_2024, yuan_streammapnet_2024} and \text{$T_\text{dist} = 5\;m$} to separate between $V_\text{close}$ and $V_\text{far}$.}
    \label{tab:model_comparison}
\end{table*}

\paragraph{Model Failure Modes}
First, we aim to analyze our proposed measures and verify our preceding hypotheses. For the following experiments, we employ MapTRv2 \cite{liao_maptrv2_2023} as a representative online mapping model. In \cref{fig:d_s_vs_M}, we evaluate per-sample performance on all validation samples from the nuScenes original split and plot the newly proposed measure $M$ versus $d(v)$ (left) and versus $s(v)$ (right). The performance drop between  $V_\text{close}$ and $V_\text{far}$ can be clearly observed with the new performance measure, as well as degrading performance for decreasing sample similarity with even stronger correlation in comparison. The same effects can be observed for all other examined splits, since the failure mode measures $\mathcal{O}_\text{loc}$ and $\mathcal{O}_\text{geom}$ in \cref{tab:split_comparison} yield positive results across all datasets and splits. Furthermore, the performance results for mAP and $M$ in \cref{tab:split_comparison} are coherent, suggesting that $M^{\pm IQR}$ is a suitable performance measure.

\cref{tab:model_comparison} compares localization overfitting ($\mathcal{O}_\text{loc}$) and geometric overfitting ($\mathcal{O}_\text{geom}$) across the previously selected models on all datasets and splits. The results indicate that all models exhibit noticeable overfitting, but the extent varies by architecture and dataset. MapQR, for instance, shows the strongest localization overfitting on nuScenes. In contrast, MapTRv2 achieves comparatively lower localization overfitting scores on Argoverse 2, suggesting stronger generalization capabilities. The results indicate that model performance is shaped by both input memorization and geometric similarity, underscoring the need for evaluation protocols that separate these effects. 

\paragraph{Dataset Bias}
In \cref{tab:split_comparison}, we list dataset split measures for geometric diversity ($\mathrm{geomdiv(T)}$ and $\mathrm{geomdiv(V)}$), geometric similarity ($\mathrm{geomsim(T, V)}$) and geographical bias ($d(v) < 5\,m$). The original splits of both datasets demonstrate significant geographical overlap, in addition to substantial similarity across the training and validation sets, suggesting the presence of a bias.

The geographical splits proposed in \cite{lilja_localization_2024} and \cite{yuan_streammapnet_2024} exhibit minimal geographical overlap and low geometric similarity between training and validation sets, leading to a significant performance drop. A detailed comparison of the two geographical splits reveals that a high level of diversity in the training set (high $\mathrm{geomdiv}(T)$) is advantageous for the model. Both splits demonstrate only minor deviations for both $\mathrm{geomdiv}(V)$ and $\mathrm{geomsim}(T,V)$. In contrast, the geographical training splits from \cite{yuan_streammapnet_2024} exhibit a higher level of diversity compared to those from \cite{lilja_localization_2024}, resulting in enhanced performance.

Evaluation on the geometric splits shows that the geometric MST-based split derivation leads to much larger $\mathrm{geomsim}(T,V)$ compared to the previous splits. The low geometric similarity between the training and validation split also leads to low geographical overlap, due to the correlation of $d(v)$ and $s(v)$ (cf. \cref{fig:d_s_correlation}). As a consequence, the performance measure $M$ suggests a decrease in performance compared to other splits for both datasets, while the mAP result for nuScenes is similar to the geographical splits. For Argoverse 2, both mAP and $M$ indicate a significant performance drop to original and geographical splits.

\begin{figure}[t] 
\includegraphics[width=0.48\textwidth]{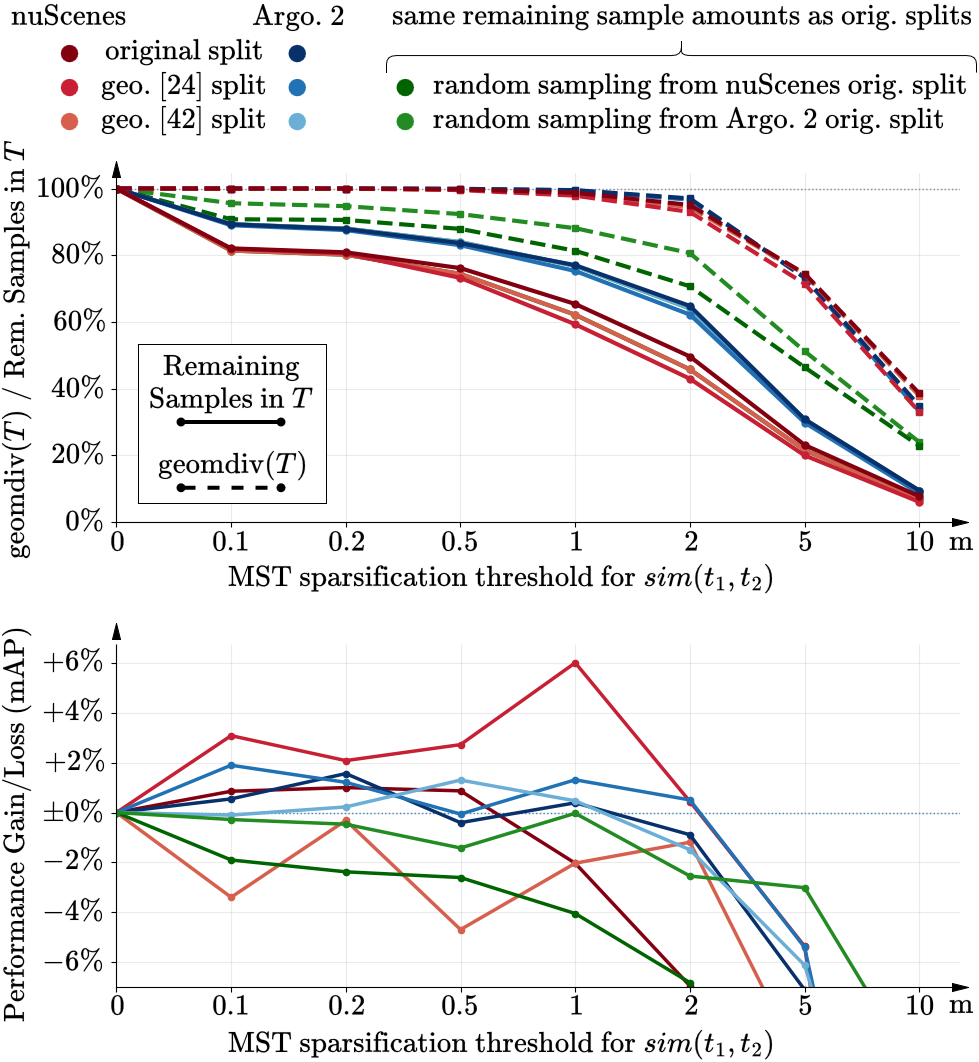}
\caption{Effect of MST-based training set sparsification on sample size and diversity (Top) and model performance for MapTRv2 on the validation set (Bottom). Besides examining the effect across all examined nuScenes and Argoverse 2 splits, we randomly sample from the training sets of the original splits for comparison.}
\label{fig:mst_sparsification_vis}
\end{figure}

\subsection{Map Geometry-Based Training Set Sparsification}

We also conduct ablation studies for the dataset diversity measure, to verify our hypothesis that a model trained on a geometrically diverse and well-balanced dataset outperforms one trained on a more homogeneous or imbalanced dataset. To demonstrate this, the MST for the training set is taken across all datasets and splits, and sparsification is performed based on a geometric similarity threshold. All samples that are connected with edges below the threshold belong to one cluster, for which a representative sample is chosen by the lowest average neighbor weight. In \cref{fig:mst_sparsification_vis} we show the remaining sample amounts, $\mathrm{geomdiv}(T)$ and  validation set performance (mAP) for the sparsified training sets. For low thresholds (0.1-1), the cumulative MST edge length $\mathrm{geomdiv}(T)$ remains practically unchanged, while the sample amounts drop significantly. For higher thresholds (2-10), both quantities decline. 
Interestingly, the performance seems to increase for most splits between sparsification thresholds 0.1 and 1, suggesting that multiple samples with very similar map geometry do not benefit the model and removing each but one leads to a more well-balanced training of the model, without overemphasizing specific map geometries. For higher thresholds (2-10) the performance decreases, because more samples are being removed leading to significant loss of information in the training data. 
To confirm that our MST-based sparsification strategy is effective, we compared the sparsified training sets from the original splits of nuScenes and Argoverse 2 with randomly sampled ones with the same remaining sample amounts. In contrast to our strategy, the geometric diversity decreases significantly even for low thresholds (0.1-1) and random sampling never leads to any performance gains. Exact numbers for \cref{fig:mst_sparsification_vis} are provided in the supplementary material.

\section{Conclusion}
\label{sec:conclusion}

We introduced a failure-mode–aware evaluation for online mapping that separates dependence on location-specific features from sensitivity to map geometry. Using distance to training data and a geometric similarity measure, we built stratified subsets to reveal whether models are memorizing features or geometric structures or truly generalizing. A Fréchet distance-based, order-aware performance measure complemented established metrics on small and imbalanced splits. We also quantified dataset geometric diversity with a MST approach and showed how it can guide pruning of redundant samples. Across two benchmarks, the analysis revealed clear signs of feature memorization and consistent performance drop with lower geometric similarity, and linked higher diversity in training data to better performance.

In future work, we aim to transfer the concept of diversity measurements through feature-based MSTs to other domains and investigate its general applicability. We also plan to explore map geometry-aware training objectives as a means to reduce geometric overfitting and to leverage diversity and similarity measures for guiding active data selection. Furthermore, we see potential in scaling similarity estimation by employing faster or learned approximations, enabling our methods to operate effectively on larger datasets.
\section*{Acknowledgements}
This work is a result of the joint research project STADT:up (19A22006B). The project is supported by the German Federal Ministry for Economic Affairs and Climate Action (BMWK), based on a decision of the German Bundestag. The authors are solely responsible for the content of this publication.
{
    \small
    \bibliographystyle{ieeenat_fullname}
    \bibliography{main}
}

\clearpage
\setcounter{page}{1}
\maketitlesupplementary

\begin{strip}
\centering
\includegraphics[width=\textwidth]{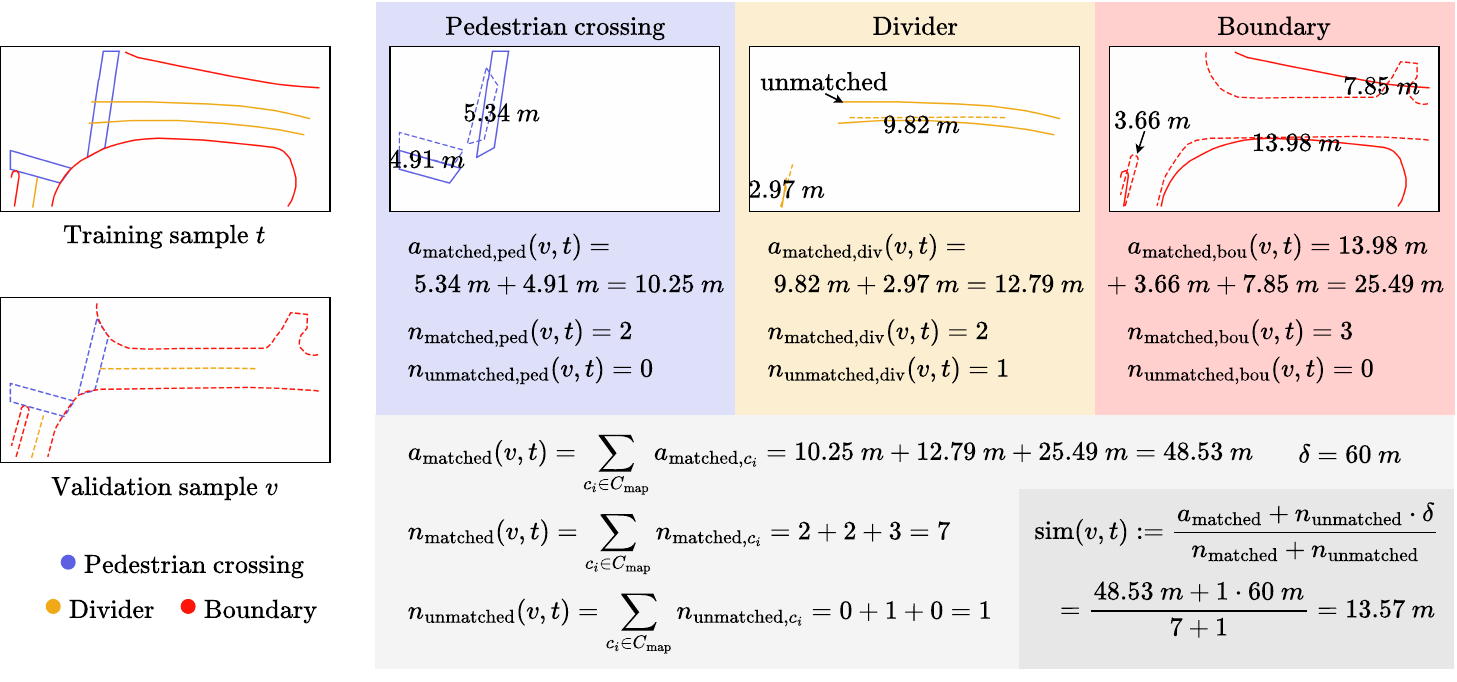}
\captionof{figure}{Visualization of an exemplary calculation of $\mathrm{sim}(v,t)$ for a sample pair $v \in V$ (dashed lines) and $t \in T$ (solid lines) from the nuScenes near extrapolation split proposed in \cite{lilja_localization_2024}. Per class, matches are found by calculating pairwise discrete Fréchet distances between map elements and bipartite matching on the resulting cost matrix. Finally, the matched and unmatched costs are summed and divided by the number of matched and unmatched elements, yielding the final value for $\mathrm{sim}(v,t)$. The two samples have no geometrical overlap.}
\label{fig:frechet_metric_vis}
\end{strip}

\section{Visual Example of Sample Similarity Computation}

Following the definition of $\mathrm{sim}(v, t)$ in \cref{eval_set_derivation}, \cref{fig:frechet_metric_vis} visualizes its computation for an exemplary sample pair from the geographically disjoint near extrapolation nuScenes split introduced in \cite{lilja_localization_2024}.

\section{Metric Definitions}

In the following, we provide the mathematical definitions for the Chamfer distance used in the AP metric in \cref{sec:performance_scores} and the discrete Fréchet distance \cite{eiter_computing_1994} used in $\mathrm{sim}(\cdot,\cdot)$ in \cref{eval_set_derivation} and $M$ in \cref{sec:performance_scores}.

\subsection{Chamfer Distance}

Let $P, Q \subset \mathbb{R}^d$ be two finite point sets, each representing one map element (polygon or polyline) after uniformly resampling it to $N_{\text{pts}}$ vertices.  
The Chamfer distance $d_{\mathrm{ch}}$ between $P$ and $Q$ is defined as
\begin{equation}
    \begin{split}
        d_{\mathrm{ch}}&(P, Q) = \\
        &\frac{1}{2}
        \left(
        \frac{1}{|P|}
        \sum_{p \in P}
        \min_{q \in Q} \|p - q\|_2
        +
        \frac{1}{|Q|}
        \sum_{q \in Q}
        \min_{p \in P} \|q - p\|_2
        \right).
    \end{split}
\end{equation}

\subsection{Discrete Fréchet Distance} \label{sec:discrete_frechet_distance}

Let $P$ and $Q$ be two map elements represented as polygons or polylines in $\mathbb{R}^d$, and let

\begin{figure*}[ht] 
\centering
\includegraphics[width=0.8\textwidth]{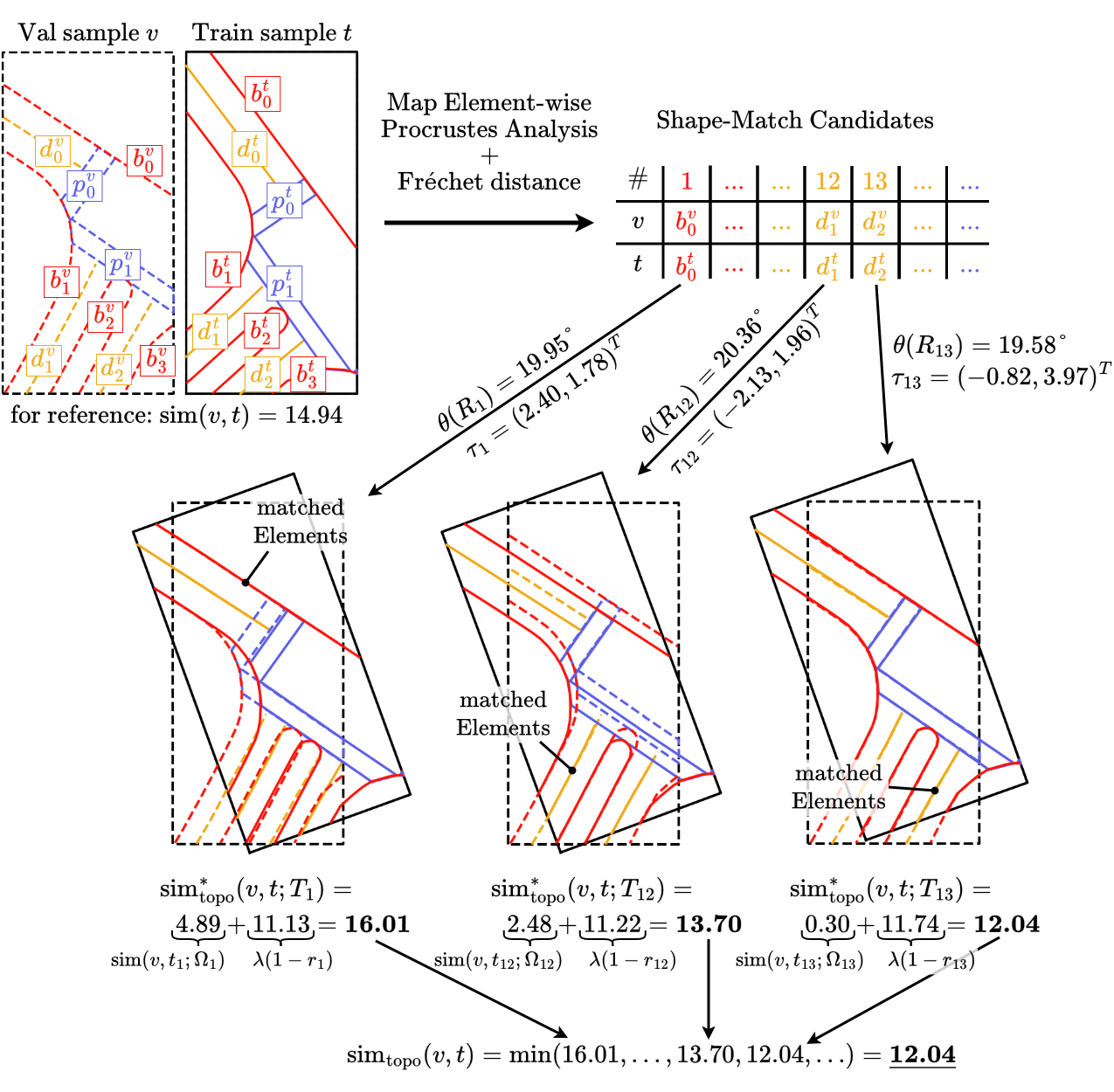}
\caption{Visualization of the topology-based similarity measure $\mathrm{sim}_{\mathrm{topo}}(v,t)$ and its computation steps. Candidate rigid transformations are derived via class and map element-wise Procrustes analysis and discrete Fréchet distance matching. For each transformed training sample, similarity is evaluated within the overlapping field-of-view and penalized for low overlap. The final similarity is obtained by selecting the minimum across all candidates.}
\label{fig:topo_alignment}
\end{figure*}

\begin{equation}
    \sigma(P) = (p_1,\dots,p_n), \qquad
    \sigma(Q) = (q_1,\dots,q_m)
\end{equation}

be the corresponding sequences of uniformly resampled vertices, with $p_i, q_j \in \mathbb{R}^d$.

A \emph{coupling} $L$ between $P$ and $Q$ is a sequence of distinct pairs between
$\sigma(P)$ and $\sigma(Q)$,

\begin{equation}
    L = \bigl( (p_{a_1}, q_{b_1}), \dots, (p_{a_K}, q_{b_K}) \bigr),
\end{equation}

where $(a_k)_{k=1}^K$ and $(b_k)_{k=1}^K$ are nondecreasing surjective index sequences, i.e.

\begin{gather}
    a_1 = 1,\; a_K = n,\; b_1 = 1,\; b_K = m, \\
    \{a_1,\dots,a_K\} = \{1,\dots,n\}, \quad
    \{b_1,\dots,b_K\} = \{1,\dots,m\},    
\end{gather}

and for all $r < s$,

\begin{equation}
    a_r \le a_s, \qquad b_r \le b_s.
\end{equation}

The norm $\|L\|$ of a coupling $L$ is the length of its longest pair,

\begin{equation}
    \|L\|
    =
    \max_{k = 1,\dots,K} \|p_{a_k} - q_{b_k}\|_2.
\end{equation}
    
The discrete Fréchet distance $d_\text{fr}$ between $P$ and $Q$ is then defined as

\begin{equation}
    d_\text{fr}(P,Q)
    =
    \min\bigl\{\|L\| \;\big|\; L \text{ is a coupling between } P \text{ and } Q\bigr\}.
\end{equation}


\begin{figure*}[ht] 
\centering
\includegraphics[width=0.9\textwidth]{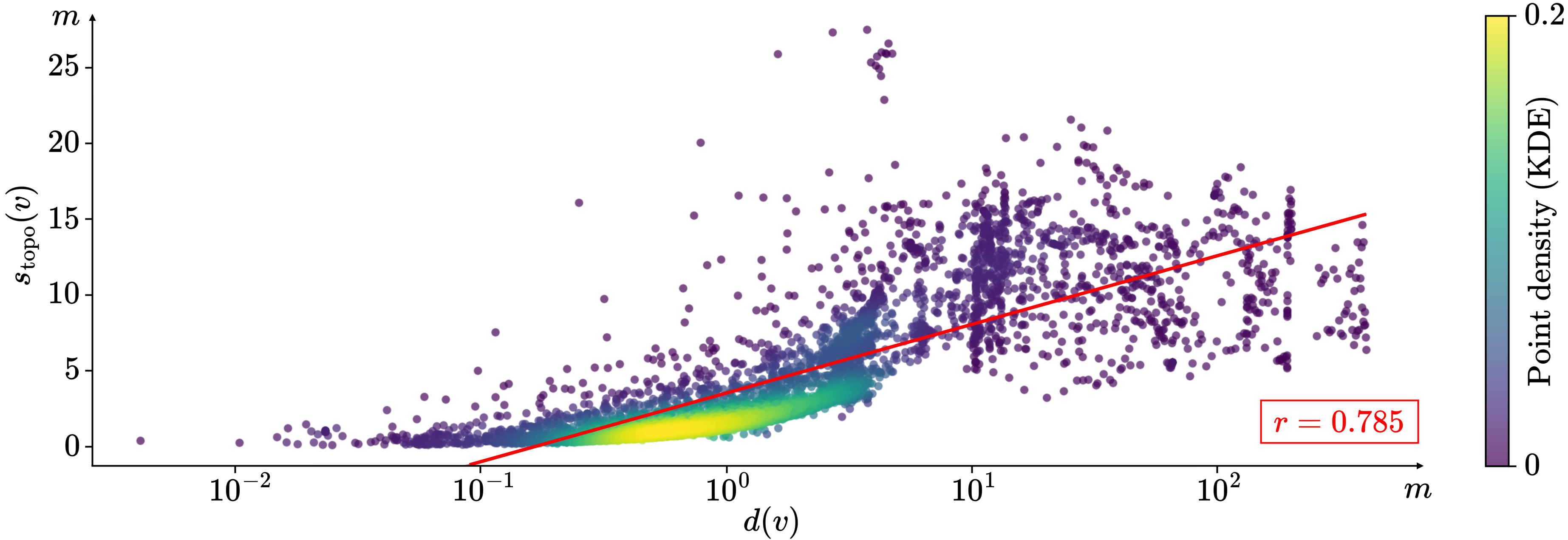}
\caption{Correlation between the topology-based similarity $s_{\mathrm{topo}}(v)$ and the geographical distance $d(v)$ for the nuScenes original split. Compared to the geometric similarity $s(v)$, the topology-based score exhibits a stronger correlation and a less cluttered distribution (Pearson correlation coefficient $r = 0.785 > 0.724$, cf. \cref{fig:d_s_correlation}).}
\label{fig:d_s_correlation_topo}
\end{figure*}

\section{Topology-Based Similarity Measure}

\paragraph{Motivation}
Besides examining the effect of geometrical similarity, we argue that online mapping models are capable of learning translation- and rotation-invariant features. We therefore introduce the similarity measure $\mathrm{sim}_\text{topo}(v, t)$ to measure translation- and rotation-invariant similarity between samples, comparing their topological structures rather than the geometrical patterns in the field of view as in $\mathrm{sim}(v, t)$. We base $\mathrm{sim}_\text{topo}(v, t)$ on $\mathrm{sim}(v, t)$ with a preceding alignment step. A visualization of the process to derive the topology-based similarity measure for an exemplary validation sample is shown in \cref{fig:topo_alignment}, we advise to follow the steps in the figure along with the mathematical definition below.

\paragraph{Definition}
Let $\Omega_v \subset \mathbb{R}^2$ and $\Omega_t \subset \mathbb{R}^2$ denote the FOV of the validation sample $v$ and the training sample $t$, respectively.
We first obtain a finite set of candidate rigid transformations

\begin{equation}
  \mathcal{T}(v,t) = \{T_k(x) = R_k x + \tau_k \mid k = 1,\dots,K\},  
\end{equation}

by comparing the shapes of all map elements within the same class using
Procrustes analysis (without uniform scaling) followed by measuring similarity using discrete Fréchet distance (cf. \cref{sec:discrete_frechet_distance}) and selecting the $k$ top matches. Each transformation $T_k$ is defined by a rotation matrix
$R_k \in \mathrm{SO}(2)$ and a translation vector $\tau_k \in \mathbb{R}^2$.

For a given candidate $T_k$, we transform the training sample

\begin{equation}
  t_k := T_k(t),
\end{equation}

and consider only the part of the scene that lies in the overlapping FOV

\begin{equation}
  \Omega_k := \Omega_v \cap T_k(\Omega_t).
\end{equation}

We denote by $\mathrm{sim}(v, t_k; \Omega_k)$ the original similarity
$\mathrm{sim}$ evaluated only on those map elements of $v$ and $t_k$ whose
geometry lies inside or intersects $\Omega_k$. All elements outside $\Omega_k$
are discarded and intersecting elements are clipped, since elements outside of $\Omega_k$ cannot occur in the opposing sample.

To avoid degenerate alignments through trivial solutions where the overlap between FOVs becomes very small and no map elements lie inside or intersect with $\Omega_k$, we penalize low-overlap candidates. We define the overlap ratio

\begin{equation}
  r_k \;:=\; \frac{\lvert \Omega_k \rvert}{\lvert \Omega_v \rvert} \in [0,1]
\end{equation}

and add a penalty term that increases linearly as the overlap decreases.
With a weight parameter $\lambda \ge 0$, we define the topology-based
similarity for a candidate transform $T_k$ as

\begin{equation}
  \mathrm{sim}^\ast_\text{topo}(v, t; T_k)
  \;:=\;
  \mathrm{sim}(v, t_k; \Omega_k)
  \;+\;
  \lambda \bigl(1 - r_k\bigr).
\end{equation}

Finally, the topology-based similarity between $v$ and $t$ is obtained by
minimizing over all candidate transforms:

\begin{equation}
  \mathrm{sim}_\text{topo}(v, t)
  \;:=\;
  \min_{T_k \in \mathcal{T}(v,t)}
  \mathrm{sim}^\ast_\text{topo}(v, t; T_k).
\end{equation}

Analogous to $s(v)$, we define $s_\text{topo}(v)$ as the lowest similarity cost between $v$ and any training sample $t \in T$ with the new topological similarity measure across all candidate transforms

\begin{equation}
    s_\text{topo}(v) := \min_{t \in T} \mathrm{sim}_\text{topo}(v, t).
\end{equation}

\paragraph{Results}
For this ablation study, we aim to validate our translation- and rotation-invariant similarity measure $s_\text{topo}$ and compare it against $s$ in terms of their correlation with performance. This allows us to determine whether overfitting is more strongly driven by geometric patterns or by topological structure.

\begin{figure*}[ht] 
\centering
\includegraphics[width=0.95\textwidth]{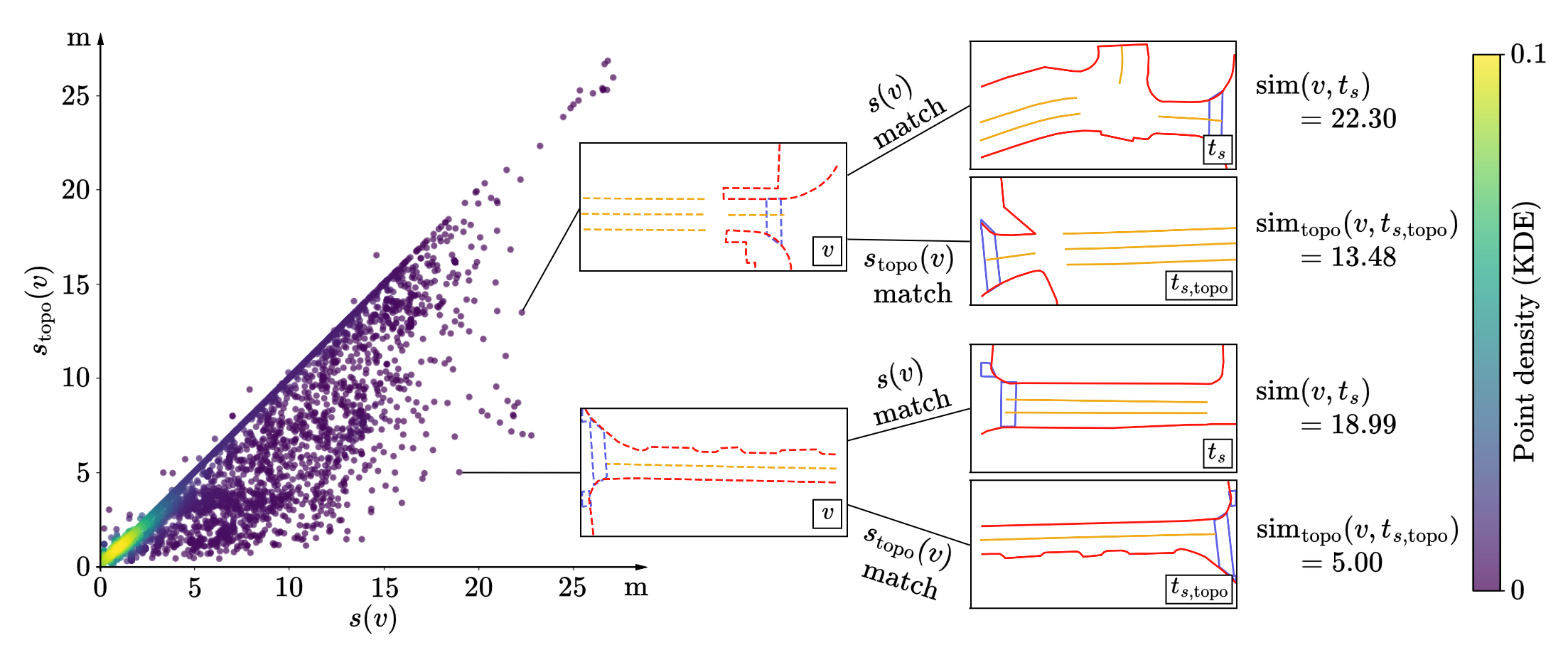}
\caption{Comparison of geometric similarity $s(v)$ and topology-based similarity $s_{\mathrm{topo}}(v)$ for the nuScenes original split. Samples deviating from the diagonal most often satisfy $s(v) > s_{\mathrm{topo}}(v)$, demonstrating that $s_{\mathrm{topo}}(v)$ identifies topologically similar scenes that differ due to translation or rotation. Two exemplary validation samples are displayed along their best matches from the training set for $s(v)$ and $s_{\mathrm{topo}}(v)$.}
\label{fig:s_vs_s_topo}
\end{figure*}

We begin by examining the correlation between $s_\text{topo}(v)$ and $d(v)$ in the nuScenes original split in \cref{fig:d_s_correlation_topo}. As expected, the correlation is stronger compared to $s(v)$, and the plot is visibly less cluttered (Pearson correlation coefficient $r = 0.785 > 0.724$, cf. \cref{fig:d_s_correlation}). This is because samples that lie close to each other tend to share similar topological structure, whereas their geometric structure, which is sensitive to translation and rotation, differs more significantly. This suggests that the metric accurately captures topological structure rather than pure geometric alignment. 
To further substantiate this claim, we compare $s(v)$ and $s_{\text{topo}}(v)$ in \cref{fig:s_vs_s_topo}. Several samples deviate from the bisector of the two axes, most often with $s(v) > s_{\text{topo}}(v)$, indicating that a closer topological match has been identified (cf. examples in \cref{fig:s_vs_s_topo}).


\begin{figure}[t] 
\includegraphics[width=0.48\textwidth]{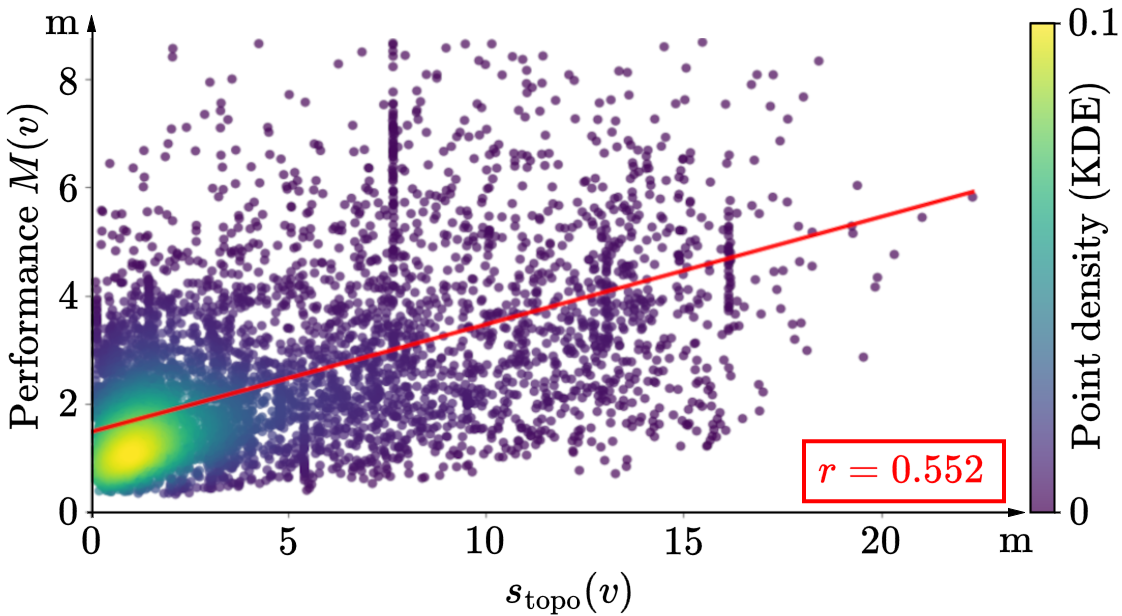}
\caption{Correlation between the topology-based similarity $s_{\mathrm{topo}}(v)$ and the per-sample performance $M(v)$ on the nuScenes original split. The slightly weaker correlation compared to $s(v)$ (Pearson correlation coefficient $r = 0.552 < 0.568$, cf. \cref{fig:d_s_vs_M}) indicates that current models rely more strongly on geometric patterns than on topological structure.}
\label{fig:s_vs_M_topo}
\end{figure}

\begin{table}[t]
    \centering
    \begin{tabular}{llcc}
        \multicolumn{2}{l}{Dataset and split} & $r(s(v), M(v))$ & $r(s_\text{topo}(v), M(v))$ \\[0.3em]
        \toprule
        \multirow{4}{*}{\rotatebox{90}{nuScenes}} & original & \textbf{0.568} & 0.552 \\
        & geo. \cite{lilja_localization_2024} & \textbf{0.226} & 0.211 \\
        & geo. \cite{yuan_streammapnet_2024} & \textbf{0.275} & 0.270 \\
        & geometric & 0.137 & \textbf{0.171} \\
        \midrule
        \multirow{4}{*}{\rotatebox{90}{Argoverse 2}} & original & \textbf{0.392} & 0.298 \\
        & geo. \cite{lilja_localization_2024} & \textbf{0.433} & 0.366 \\
        & geo. \cite{yuan_streammapnet_2024} & \textbf{0.330} & 0.202 \\
        & geometric & -0.071 & \textbf{0.209} \\
    \end{tabular}
    \caption{Pearson correlation between $M(v)$ and geometric similarity $s(v)$ versus topology-based similarity $s_{\mathrm{topo}}(v)$ across all examined dataset splits. While $s(v)$ correlates more strongly with performance for original and geographical splits, $s_{\mathrm{topo}}(v)$ shows mildly higher correlation for the geometric splits, where geometric similarity is intentionally minimized.}
    \label{tab:s_M_correlation_comparison_geom_topo}
\end{table}

We also reexamine the correlation between $s_\text{topo}(v)$ and the per-sample performance $M(v)$ on the original nuScenes split in \cref{fig:s_vs_M_topo}. The Pearson correlation is slightly lower for $s_{\text{topo}}(v)$ than for $s(v)$ ($r = 0.552 < 0.568$, cf. \cref{fig:d_s_vs_M}), suggesting that the online mapping model relies more on rotation- and translation-dependent geometric features than on invariant topological features.

To support this claim, we examine the correlation between $s(v)$ and $M(v)$ against the correlation between $s_\text{topo}(v)$ and $M(v)$ for all splits that are examined in \cref{sec:experiments}. The results are shown in \cref{tab:s_M_correlation_comparison_geom_topo}. While the Pearson correlation is in a similar range per split, we see slightly stronger correlation for the geometrical similarity measure $s(v)$ across all original and geographical splits. However, in the geometrical splits we introduced, the correlation for $s_\text{topo}$ is higher, even though both values indicate minimal correlation to $M$. This indicates, that for these validation sets where geometries are contained that are minimize similarity to the training split, topological alignment is still beneficial for model performance.

Due to the lack of significant differences in the correlation, we expect similar results for any measures derived from $s_\text{topo}(v)$ in place of $s(v)$. Since the correlation with performance for all original and geographical splits is lower for $s_\text{topo}(v)$ compared to $s(v)$ and the effect of benefiting from topological alignment only seems to show for sets with very dissimilar geometries, we refrain from additional experiments.

\section{Details on the Geometric Similarity-Based Performance Measure}

In the following, we provide additional details on $M$ which is defined in \cref{sec:performance_scores}. Our objective is to offer a clearer intuition for $M$ and justify why it provides a more faithful assessment of geometric reconstruction quality than traditional AP-style metrics.

We obtain class-wise bipartite matching based on the discrete Fréchet distance (cf. \cref{sec:discrete_frechet_distance}) between the prediction map elements $p_{c,i} \in P_{c}$ and ground truth $g_{c,i} \in G_{c}$ to find the closest prediction for each ground truth element. In contrast to the Chamfer distance, the discrete Fréchet distance is sensitive to point ordering (cf. \cref{fig:chamfer_frechet_comparison}). For all examined online mapping architectures in the experiments \cite{liao_maptrv2_2023, liao_maptr_2023, liu_leveraging_2024, liu_mgmap_2024}, the predicted map elements have the same number of points compared to the ground truth. If the number of vertices per map element do not match, we advise to subsample or simplify to predicted map elements w.r.t. to the number of vertices in the ground truth, since the discrete Fréchet distance comparison between prediction and ground truth relies on discrete points and large distances between points could lead to inaccurate results.

Note that similarly to the Chamfer-based AP metric, multiple predictions for the same ground truth element are not penalized in order to not introduce overwhelming complexity to the metric. In practice, we examine that the predicted map elements that are matched to a ground truth map element yield significantly higher confidence scores in comparison to the unmatched predictions, so the proposed measure could also be used to determine a suitable confidence threshold for deployment of a model.

The visualization in \cref{fig:frechet_metric_vis} illustrates how the matching between two samples works for $\mathrm{sim}(v, t)$, the same concept also applies to the measure $M$ but instead of comparing two samples, we compare the sets of predictions and ground truth. Furthermore, instead of summing the matched costs resulting in $a_\text{matched}$, we aggregate all matched costs into a distribution $D$ that represents the map-element-wise geometric reconstruction quality. We then take the median $M$ and interquartile range $IQR$ to characterize $D$ in a comparable way, which remains informative for sets of varying sizes, being less sensitive to discrete outcomes in comparison to the Chamfer distance-based AP metric and enabling comparisons at the single-sample level.

\section{Geometrical Dataset Split Derivation} \label{sec:geom_split_deriv}

Besides the original and geographical training splits, we want to examine model performance on a geometric dataset split, focusing on maximum geometric dissimilarity between training and evaluation sets. To be coherent with the other splits, the geometric split should also partition the data into 70/15/15\% for the training, validation, and test sets, respectively. 

We base our geometric split on the geometric similarity MST for the whole dataset. At first, we identify the edges in the MST with highest similarity costs, suggesting highest geometric dissimilarity. We then consider these edges as first-cut candidates and retain those that yield a subset of size near the desired training set size (70\% of the dataset size). For each retained first cut, we search for a second cut in the remaining opposite subset (30\% of the dataset size) to separate it into the validation and test set (each 15\% of the dataset size). We then evaluate the resulting three subsets against size tolerances, and score candidates by a balance–separation criterion (small size deviation, large cut edge weights). Finally, we remove the chosen two edges and assign the three connected subtrees to train/val/test sets.

\section{Exact Numbers for Dataset Sparsification}\label{sec:exact_numbers}

In the following, we list the exact numbers for \cref{fig:mst_sparsification_vis} for the reader to reference. For each split, we report the sparsification threshold, remaining samples, MST length, and mAP values on the validation set corresponding to the plotted results.

\begin{table*}[h]
\centering
\begin{tabular}{r@{\hspace{15pt}}l@{\hspace{17pt}}r@{\hspace{5pt}}l@{\hspace{12pt}}l@{\hspace{17pt}}r@{\hspace{5pt}}l@{\hspace{12pt}}l@{\hspace{17pt}}r@{\hspace{5pt}}l}
\toprule
Sparsification Threshold & \multicolumn{3}{c}{Remaining Samples in $T$} & \multicolumn{3}{c}{$\mathrm{geomdiv}(T)$} & \multicolumn{3}{c}{Performance on $V$ (mAP)} \\
\midrule
--  & & 28130 & (100.00 \%) & & 96828.00 m & (100.00 \%) & & 60.95 & ($\pm$0.00 \%) \\
0.1 & & 23084 & (82.06 \%) & & 96815.33 m & (99.99 \%) & & 61.47 & (+0.85 \%) \\
0.2 & & 22752 & (80.88 \%) & & 96805.15 m & (99.98 \%) & & 61.56 & (+1.00 \%) \\
0.5 & & 21417 & (76.14 \%) & & 96613.71 m & (99.78 \%) & & 61.48 & (+0.87 \%) \\
1   & & 18397 & (65.40 \%) & & 95585.00 m & (98.72 \%) & & 59.71 & (-2.03 \%) \\
2   & & 13969 & (49.66 \%) & & 92051.27 m & (95.07 \%) & & 56.71 & (-6.96 \%) \\
5   & & 6512  & (23.15 \%) & & 72116.01 m & (74.48 \%) & & 48.11 & (-21.07 \%) \\
10  & & 2184  & (7.76 \%) & & 37363.45 m & (38.59 \%) & & 28.50 & (-53.24 \%) \\
\bottomrule
\end{tabular}
\caption{Original nuScenes split}
\label{tab:orig_nusc_sparisifcation}
\end{table*}

\begin{table*}[h]
\centering
\label{tab:geo_local_nusc_sparisifcation}
\begin{tabular}{r@{\hspace{15pt}}l@{\hspace{17pt}}r@{\hspace{5pt}}l@{\hspace{12pt}}l@{\hspace{17pt}}r@{\hspace{5pt}}l@{\hspace{12pt}}l@{\hspace{17pt}}r@{\hspace{5pt}}l}
\toprule
Sparsification Threshold & \multicolumn{3}{c}{Remaining Samples in $T$} & \multicolumn{3}{c}{$\mathrm{geomdiv}(T)$} & \multicolumn{3}{c}{Performance on $V$ (mAP)} \\
\midrule
--  & & 27840 & (100.00 \%) & & 80572.21 m & (100.00 \%) & & 24.96 & ($\pm$0.00 \%) \\
0.1 & & 22815 & (81.95 \%) & & 80559.35 m & (99.98 \%) & & 25.73 & (+3.08 \%) \\
0.2 & & 22427 & (80.56 \%) & & 80543.07 m & (99.96 \%) & & 25.48 & (+2.08 \%) \\
0.5 & & 20381 & (73.21 \%) & & 80197.95 m & (99.54 \%) & & 25.64 & (+2.72 \%) \\
1   & & 16515 & (59.32 \%) & & 78843.58 m & (97.85 \%) & & 26.46 & (+6.01 \%) \\
2   & & 11970 & (43.00 \%) & & 74909.77 m & (92.97 \%) & & 25.07 & (+0.44 \%) \\
5   & & 5575  & (20.03 \%) & & 57577.21 m & (71.46 \%) & & 23.62 & (-5.37 \%) \\
10  & & 1697  & (6.10 \%)  & & 26597.42 m & (33.01 \%) & & 15.59 & (-37.54 \%) \\
\bottomrule
\end{tabular}
\caption{Geographically disjoint nuScenes split from \cite{lilja_localization_2024} (Near Extrapolation Split)}
\end{table*}

\begin{table*}[h]
\centering
\label{tab:geo_stream_nusc_sparisifcation}
\begin{tabular}{r@{\hspace{15pt}}l@{\hspace{17pt}}r@{\hspace{5pt}}l@{\hspace{12pt}}l@{\hspace{17pt}}r@{\hspace{5pt}}l@{\hspace{12pt}}l@{\hspace{17pt}}r@{\hspace{5pt}}l}
\toprule
Sparsification Threshold & \multicolumn{3}{c}{Remaining Samples in $T$} & \multicolumn{3}{c}{$\mathrm{geomdiv}(T)$} & \multicolumn{3}{c}{Performance on $V$ (mAP)} \\
\midrule
--  & & 28008 & (100.00 \%) & & 90172.57 m & (100.00 \%) & & 28.53 & ($\pm$0.00 \%) \\
0.1 & & 22801 & (81.41 \%) & & 90159.43 m & (99.99 \%) & & 27.56 & (-3.40 \%) \\
0.2 & & 22451 & (80.16 \%) & & 90148.35 m & (99.97 \%) & & 28.44 & (-0.32 \%) \\
0.5 & & 20840 & (74.41 \%) & & 89897.52 m & (99.69 \%) & & 27.19 & (-4.70 \%) \\
1   & & 17414 & (62.18 \%) & & 88718.11 m & (98.39 \%) & & 27.95 & (-2.03 \%) \\
2   & & 12825 & (45.79 \%) & & 85090.37 m & (94.36 \%) & & 28.19 & (-1.19 \%) \\
5   & & 6051  & (21.60 \%) & & 67000.17 m & (74.30 \%) & & 25.57 & (-10.38 \%) \\
10  & & 1964  & (7.01 \%)  & & 34123.62 m & (37.84 \%) & & 18.10 & (-36.56 \%) \\
\bottomrule
\end{tabular}
\caption{Geographically disjoint nuScenes split from \cite{yuan_streammapnet_2024}}
\end{table*}

\begin{table*}[h]
\centering
\begin{tabular}{r@{\hspace{15pt}}l@{\hspace{17pt}}r@{\hspace{5pt}}l@{\hspace{12pt}}l@{\hspace{17pt}}r@{\hspace{5pt}}l@{\hspace{12pt}}l@{\hspace{17pt}}r@{\hspace{5pt}}l}
\toprule
Sparsification Threshold & \multicolumn{3}{c}{Remaining Samples in $T$} & \multicolumn{3}{c}{$\mathrm{geomdiv}(T)$} & \multicolumn{3}{c}{Performance on $V$ (mAP)} \\
\midrule
--  & & 22279 & (100.00 \%) & & 91003.65 m & (100.00 \%) & & 63.97 & ($\pm$0.00 \%) \\
0.1 & & 19892 & (89.29 \%) & & 90995.80 m & (99.99 \%) & & 64.32 & (+0.55 \%) \\
0.2 & & 19580 & (87.89 \%) & & 90991.00 m & (99.99 \%) & & 64.97 & (+1.56 \%) \\
0.5 & & 18633 & (83.63 \%) & & 90909.49 m & (99.90 \%) & & 63.71 & (-0.41 \%) \\
1   & & 17159 & (77.02 \%) & & 90555.71 m & (99.51 \%) & & 64.22 & (+0.39 \%) \\
2   & & 14440 & (64.81 \%) & & 88289.72 m & (97.02 \%) & & 63.40 & (-0.89 \%) \\
5   & & 6882  & (30.89 \%) & & 66745.42 m & (73.34 \%) & & 59.42 & (-7.11 \%) \\
10  & & 2116  & (9.50 \%)  & & 31739.04 m & (34.88 \%) & & 45.72 & (-28.53 \%) \\
\bottomrule
\end{tabular}
\caption{Original Argoverse 2 split, remaining samples in $T$ and $\mathrm{geomdiv}(T)$ are computed for 2 Hz to be comparable to nuScenes. Performance results are for 10 Hz.}
\label{tab:orig_av2_sparisifcation}
\end{table*}

\begin{table*}[h]
\centering
\begin{tabular}{r@{\hspace{15pt}}l@{\hspace{17pt}}r@{\hspace{5pt}}l@{\hspace{12pt}}l@{\hspace{17pt}}r@{\hspace{5pt}}l@{\hspace{12pt}}l@{\hspace{17pt}}r@{\hspace{5pt}}l}
\toprule
Sparsification Threshold & \multicolumn{3}{c}{Remaining Samples in $T$} & \multicolumn{3}{c}{$\mathrm{geomdiv}(T)$} & \multicolumn{3}{c}{Performance on $V$ (mAP)} \\
\midrule
--  & & 22223 & (100.00 \%) & & 87302.62 m & (100.00 \%) & & 49.53 & ($\pm$0.00 \%) \\
0.1 & & 19777 & (88.99 \%) & & 87294.12 m & (99.99 \%) & & 50.47 & (+1.90 \%) \\
0.2 & & 19454 & (87.54 \%) & & 87289.24 m & (99.98 \%) & & 50.13 & (+1.21 \%) \\
0.5 & & 18445 & (83.00 \%) & & 87188.20 m & (99.87 \%) & & 49.50 & (-0.06 \%) \\
1   & & 16731 & (75.29 \%) & & 86747.95 m & (99.36 \%) & & 50.18 & (+1.31 \%) \\
2   & & 13827 & (62.22 \%) & & 84389.69 m & (96.66 \%) & & 49.78 & (+0.50 \%) \\
5   & & 6636  & (29.86 \%) & & 63978.80 m & (73.28 \%) & & 46.85 & (-5.41 \%) \\
10  & & 1919  & (8.64 \%)  & & 28972.13 m & (33.19 \%) & & 36.35 & (-26.61 \%) \\
\bottomrule
\end{tabular}
\label{tab:geo_local_av2_sparisifcation}
\caption{Geographically disjoint Argoverse 2 split from \cite{lilja_localization_2024} (Near Extrapolation Split), remaining samples in $T$ and $\mathrm{geomdiv}(T)$ are computed for 2 Hz to be comparable to nuScenes. Performance results are for 10 Hz.}
\end{table*}

\begin{table*}[h]
\centering
\label{tab:geo_stream_av2_sparisifcation}
\begin{tabular}{r@{\hspace{15pt}}l@{\hspace{17pt}}r@{\hspace{5pt}}l@{\hspace{12pt}}l@{\hspace{17pt}}r@{\hspace{5pt}}l@{\hspace{12pt}}l@{\hspace{17pt}}r@{\hspace{5pt}}l}
\toprule
Sparsification Threshold & \multicolumn{3}{c}{Remaining Samples in $T$} & \multicolumn{3}{c}{$\mathrm{geomdiv}(T)$} & \multicolumn{3}{c}{Performance on $V$ (mAP)} \\
\midrule
--  & & 23954 & (100.00 \%) & & 97153.16 m & (100.00 \%) & & 57.61 & ($\pm$0.00 \%) \\
0.1 & & 21391 & (89.30 \%) & & 97144.47 m & (99.99 \%) & & 57.55 & (-0.10 \%) \\
0.2 & & 21077 & (87.99 \%) & & 97138.62 m & (99.99 \%) & & 57.74 & (+0.23 \%) \\
0.5 & & 20106 & (83.94 \%) & & 97050.71 m & (99.89 \%) & & 58.36 & (+1.30 \%) \\
1   & & 18403 & (76.83 \%) & & 96625.41 m & (99.46 \%) & & 57.88 & (+0.47 \%) \\
2   & & 15348 & (64.07 \%) & & 93994.73 m & (96.75 \%) & & 56.75 & (-1.49 \%) \\
5   & & 7353  & (30.70 \%) & & 71206.04 m & (73.29 \%) & & 54.08 & (-6.13 \%) \\
10  & & 2251  & (9.40 \%)  & & 33621.14 m & (34.61 \%) & & 45.45 & (-21.11 \%) \\
\bottomrule
\end{tabular}
\caption{Geographically disjoint Argoverse 2 split from \cite{yuan_streammapnet_2024}, remaining samples in $T$ and $\mathrm{geomdiv}(T)$ are computed for 2 Hz to be comparable to nuScenes. Performance results are for 10 Hz.}
\end{table*}

\begin{table*}[h]
\centering
\label{tab:random_nusc_sparisifcation}
\begin{tabular}{l@{\hspace{15pt}}r@{\hspace{5pt}}l@{\hspace{20pt}}l@{\hspace{20pt}}r@{\hspace{5pt}}l@{\hspace{20pt}}l@{\hspace{20pt}}r@{\hspace{5pt}}l}
\toprule
\multicolumn{3}{c}{Remaining Samples in $T$} & \multicolumn{3}{c}{$\mathrm{geomdiv}(T)$} & \multicolumn{3}{c}{Performance on $V$ (mAP)} \\
\midrule
& 28130 & (100.00 \%) & & 96828.00 m & (100.00 \%) & & 60.95 & ($\pm$0.00 \%) \\
& 23084 & (82.06 \%)  & & 87906.66 m & (90.79 \%) & & 59.79 & (-1.90 \%) \\
& 22752 & (80.88 \%)  & & 87715.78 m & (90.59 \%) & & 59.50 & (-2.38 \%) \\
& 21417 & (76.14 \%)  & & 85104.96 m & (87.89 \%) & & 59.36 & (-2.61 \%) \\
& 18397 & (65.40 \%)  & & 78757.05 m & (81.34 \%) & & 58.48 & (-4.05 \%) \\
& 13969 & (49.66 \%)  & & 68450.78 m & (70.69 \%) & & 56.79 & (-6.83 \%) \\
& 6512  & (23.15 \%)  & & 44951.67 m & (46.42 \%) & & 48.79 & (-19.95 \%) \\
& 2184  & (7.76 \%)   & & 22131.76 m & (22.86 \%) & & 31.70 & (-47.99 \%) \\
\bottomrule
\end{tabular}
\caption{Random sampling for remaining sample amounts from \cref{tab:orig_nusc_sparisifcation} from the original nuScenes split}
\end{table*}

\begin{table*}[h]
\centering
\label{tab:random_av2_sparisifcation}
\begin{tabular}{l@{\hspace{15pt}}r@{\hspace{5pt}}l@{\hspace{20pt}}l@{\hspace{20pt}}r@{\hspace{5pt}}l@{\hspace{20pt}}l@{\hspace{20pt}}r@{\hspace{5pt}}l}
\toprule
\multicolumn{3}{c}{Remaining Samples in $T$} & \multicolumn{3}{c}{$\mathrm{geomdiv}(T)$} & \multicolumn{3}{c}{Performance on $V$ (mAP)} \\
\midrule
& 22279 & (100.00 \%) & & 91003.65 m & (100.00 \%) & & 63.97 & ($\pm$0.00 \%) \\
& 19892 & (89.29 \%)  & & 86989.97 m & (95.59 \%) & & 63.79 & (-0.28 \%) \\
& 19580 & (87.89 \%)  & & 86184.47 m & (94.70 \%) & & 63.67 & (-0.47 \%) \\
& 18633 & (83.63 \%)  & & 84021.93 m & (92.33 \%) & & 63.06 & (-1.42 \%) \\
& 17159 & (77.02 \%)  & & 80187.60 m & (88.11 \%) & & 63.95 & (-0.03 \%) \\
& 14440 & (64.81 \%)  & & 73394.62 m & (80.65 \%) & & 62.34 & (-2.55 \%) \\
& 6882  & (30.89 \%)  & & 46629.68 m & (51.24 \%) & & 62.04 & (-3.02 \%) \\
& 2116  & (9.50 \%)   & & 21827.41 m & (23.98 \%) & & 57.18 & (-10.61 \%) \\
\bottomrule
\end{tabular}
\caption{Random sampling for remaining sample amounts from  \cref{tab:orig_av2_sparisifcation} from the original Argoverse 2 split, remaining samples in $T$ and $\mathrm{geomdiv}(T)$ are computed for 2 Hz to be comparable to nuScenes. Performance results are for 10 Hz.}
\end{table*}

\end{document}